\documentclass[nohyperref]{article}

\usepackage{microtype}
\usepackage{subfigure}
\usepackage{graphicx}
\usepackage{capt-of}

\usepackage{amsmath,amssymb,amsthm,amsfonts,dsfont,pifont,bm,bbm,mathrsfs,mathtools}
\usepackage{booktabs,multirow,adjustbox}
\usepackage{xspace}
\usepackage[pagebackref,breaklinks,colorlinks]{hyperref}
\usepackage{cleveref}
\crefname{section}{Sec.}{Secs.}
\Crefname{section}{Section}{Sections}
\crefname{table}{Tab.}{Tabs.}
\Crefname{table}{Table}{Tables}
\crefname{figure}{Fig.}{Figs.}
\Crefname{figure}{Figure}{Figures}
\crefname{equation}{Eq.}{Eqs.}
\Crefname{equation}{Equation}{Equations}

\theoremstyle{plain}

\theoremstyle{definition}

\theoremstyle{remark}

\newcommand{\n}{{\bm n}}
\newcommand{\z}{{\bm z}}
\newcommand{\Z}{\mathcal{Z}}
\newcommand{\x}{{\bm x}}
\newcommand{\X}{\mathcal{X}}

\newcommand{\J}{{\bm J}}
\newcommand{\I}{{\bm I}}
\newcommand{\U}{{\bm U}}
\newcommand{\V}{{\bm V}}
\newcommand{\D}{{\bm D}}

\newcommand{\mL}{{\bm L}}

\DeclareMathOperator*{\argmax}{arg\,max}
\DeclareMathOperator*{\argmin}{arg\,min}

\newcommand{\FID}{\textbf{FID$\downarrow$}}  % Notation of FID score in Table.
\newcommand{\MSE}{\textbf{MSE$\downarrow$}}  % Notation of MSE score in Table.
\newcommand{\ID}{\textbf{ID$\uparrow$}}  % Notation of ID score in Table.

\newcommand{\tocite}[1]{\textcolor{red}{[TO CITE]}}

\newcommand{\method}{ReSeFa\xspace}
\usepackage{xcolor}
\definecolor{myMagenta}{rgb}{0.9,0,0.4}

\usepackage[accepted]{icml2022}

\newcommand{\titlename}{Region-Based Semantic Factorization in GANs}
\icmltitlerunning{\titlename}

\begin{document}

\icmlsetsymbol{coa}{$\dagger$}
\icmlsetsymbol{co-coa}{$\ddagger$}

%%%% Title and authors.
\twocolumn[
  \icmltitle{\titlename}
  \icmlsetsymbol{equal}{*}
  
  \begin{icmlauthorlist}
    \icmlauthor{Jiapeng Zhu}{HKUST}
    \icmlauthor{Yujun Shen}{ByteDance}
    \icmlauthor{Yinghao Xu}{CUHK}
    \icmlauthor{Deli Zhao}{Ant}
    \icmlauthor{Qifeng Chen}{HKUST}
  \end{icmlauthorlist}
  
  \icmlaffiliation{HKUST}{Department of CSE, The Hong Kong University of Science and Technology, Hong Kong, China.}
  \icmlaffiliation{ByteDance}{ByteDance, Beijing, China}
  \icmlaffiliation{CUHK}{Department of IE, The Chinese University of Hong Kong, Hong Kong, China.}
  \icmlaffiliation{Ant}{Ant Research, Hangzhou, China}

  \icmlcorrespondingauthor{Qifeng Chen}{cqf@ust.hk}
  \icmlkeywords{Machine Learning, ICML}
  \vskip 0.3in
]

\begin{NoHyper}
\printAffiliationsAndNotice{}
\end{NoHyper}

\begin{abstract}
Despite the rapid advancement of semantic discovery in the latent space of Generative Adversarial Networks (GANs), existing approaches either are limited to finding global attributes or rely on a number of segmentation masks to identify local attributes.
In this work, we present a highly efficient algorithm to factorize the latent semantics learned by GANs concerning an \textit{arbitrary} image region.
Concretely, we revisit the task of local manipulation with pre-trained GANs and formulate region-based semantic discovery as a dual optimization problem.
Through an appropriately defined generalized Rayleigh quotient, we manage to solve such a problem \textit{without any annotations or training}.
Experimental results on various state-of-the-art GAN models demonstrate the effectiveness of our approach, as well as its superiority over prior arts regarding precise control, region robustness, speed of implementation, and simplicity of use.
Our source code can be found at \href{https://github.com/zhujiapeng/resefa/}{here}.
\end{abstract}

\section{Introduction}\label{sec:intro}

Recent studies have shown that versatile semantics emerge in the latent space of pre-trained Generative Adversarial Networks (GANs)~\cite{goetschalckx2019ganalyze, shen2020interpreting, gansteerability, yang2021semantic}.
Identifying these variation factors, which are typically devised as some directions in the latent spaces~\cite{shen2020interpreting}, facilitates a wide range of downstream tasks~\cite{xu2021generative, zhang2021datasetgan, tan2020improving}, especially image editing~\cite{gansteerability, yang2021semantic, menon2020pulse, gu2020image, zhu2020domain, ling2021editgan}.
In particular, moving latent codes along a certain direction can cause corresponding semantic changes in the synthesized images.
Accordingly, it is of great use to discover these steerable directions diversely and precisely.

To interpret the latent space learned by GANs, many attempts have been made, including both supervised ones~\cite{shen2020interfacegan, gansteerability, yang2021semantic} and unsupervised ones~\cite{shen2021closed, voynov2020unsupervised, ganspace}.
Most prior arts, however, target at finding global attributes~\cite{shen2020interfacegan, yang2021semantic, voynov2020unsupervised, ganspace} such that altering the latent code with these attributes will manipulate the output image as a whole.
Researchers have given recent attention to detecting local semantics due to their more practical usage, but they usually require a number of images labeled with segmentation masks for the discovery process~\cite{suzuki2018spatially, collins2020editing, wu2020stylespace, ling2021editgan}.
A very recent work manages to relate a local image region to a GAN latent subspace independent of annotations~\cite{zhu2021lowrankgan}, yet it turns to depend on some sensitive hyper-parameters, resulting in insufficient robustness to the selected region.

In this work, we propose a surprisingly simple algorithm, termed as \method, for region-based semantic factorization in GANs.
Unlike existing methods that only treat image editing as an application and take no account of the manipulation model for semantic exploration, we re-examine the task of local editing using pre-trained GANs as the prior.
Specifically, given a region of interest, a robust manipulation method should take effect on the contents within this area only and preserve the remaining contents as much as possible.
In other words, after altering the latent code, we expect the pixels located in the target region to change while the outside pixels remain the same.
Such an analysis helps define an optimization problem based on the derivative of pixel values with respect to the latent code (\textit{i.e.}, Jacobian).
Solving this problem can help identify the variation factors corresponding to a particular image region.
We further notice that the optimization objective can be formulated as a generalized Rayleigh quotient~\cite{Horn2012matrix} such that the aforementioned problem can be solved efficiently.

To summarize, our proposed algorithm has the following advantages over prior work.
First, our algorithm does not rely on detailed spatial masks. % in a completely unsupervised manner.
Taking mouth editing with a face synthesis GAN model as an instance, \method only needs a rough bounding box around the mouth of a single synthesized image, making it sufficiently easy to use in practice.
Second, our method is purely based on solving an eigen-decomposition problem, which is independent of any hyper-parameters or model structures. % derivative-deduced has no clear meaning
Consequently, it is fairly flexible so that users can customize the regions of their interests arbitrarily with any pre-trained GAN model.
Third, thanks to the adequately defined generalized Rayleigh quotient, our approach enables a fast implementation, especially when the latent space is in high dimensions (\textit{e.g.}, $\mathcal{W}^+$ space of StyleGAN~\cite{abdal2020image2stylegan++}).
%
% Such a property enables an interactive interface, which allows users to verify their thoughts very quickly.
%
Extensive experimental results suggest that our \method shows precise controllability and strong robustness to the selected image local region, while it can be easily generalized to state-of-the-art GAN variants, including StyleGAN2~\cite{stylegan2} and BigGAN~\cite{biggan}.

\section{Related Work}

\textbf{Generative Adversarial Networks.}
GANs~\cite{gan} have significantly advanced high-fidelity image synthesis with different objective functions~\cite{wgan}, novel training schedules~\cite{pggan, biggan}, carefully designed network architectures~\cite{stylegan, stylegan2, stylegan3}, and improved data efficiency~\cite{zhao2020differentiable, stylegan2ada, yang2021data}.
Through properly reusing the knowledge learned in the GAN pre-training, prior arts have demonstrated a wide range of downstream applications of GANs, such as image classification~\cite{xu2021generative}, image segmentation~\cite{zhang2021datasetgan}, visual alignment~\cite{peebles2021gan}, image editing~\cite{gu2020image, menon2020pulse}, \textit{etc.}

\textbf{Local Editing with GANs.}
Among all the applications of GANs, image local editing earns a number of audiences considering its interactivity and practical usage.
One straightforward way of controlling the synthesis of a certain image region is to make the GAN generator spatially aware during training~\cite{lee2020maskgan, kim2021exploiting}.
An alternative way is to first segment the synthesis results and then manipulate (\textit{e.g.}, swap) the intermediate feature maps at the region of interest~\cite{suzuki2018spatially, bau2020ganpaint, collins2020editing, zhang2021decorating}.
However, all these approaches tend to perform editing only from the instance level instead of the semantic level.
Taking face local manipulation as an example, these methods are capable of harmonizing the eyes of one person to another~\cite{lee2020maskgan, kim2021exploiting, suzuki2018spatially, collins2020editing} yet fail to make a person close the eyes.
Meanwhile, they require users to specify spatial masks for each editing (\textit{e.g.}, the eyes of a person may not always locate at the same spatial position in different images), making them hard to generalize to all samples.

\textbf{Semantic Discovery in GANs.}
Interpreting the generation mechanism of GANs helps us understand the rules about how the generator renders an image.
In this way, we can utilize such rules for image editing once for all~\cite{bau2019gandissection, bau2020rewriting}.
A typical way to control the GAN generation is to identify some steerable directions within the latent space~\cite{gansteerability}.
These latent directions usually correspond to some high-level semantics, like the age of a person, and can be faithfully used for attribute manipulation of any synthesized image~\cite{goetschalckx2019ganalyze, plumerault2020controlling, shen2020interfacegan, yang2021semantic, voynov2020unsupervised, ganspace, shen2021closed, spingarn2021gan, cherepkov2021navigate, he2021eigengan}.
Nevertheless, most directions found by previous methods are targeted at the entire image, and how to discover the semantics for some image regions remains unsolved.

It has recently been shown that some latent subspaces of GANs can be directly used for image local editing without operating the feature maps~\cite{wu2020stylespace, lang2021explaining, zhu2021lowrankgan, ling2021editgan}.
\citet{wu2020stylespace} propose StyleSpace, which uncovers the relationship between some convolutional units in the generator and the objects within the output image,
however, identifying the object-oriented channels requires a number of object masks as the ground-truth and is only applicable to style-based network structure~\cite{stylegan}.
\citet{ling2021editgan} propose a novel local editing approach by controlling the segmentation mask.
However, the manipulation pipeline requires manually editing the segmentation mask, which needs skilled personnel and precise ground truth, and requires optimizing the latent code to meet the expected semantic change, which can be time-consuming.
In addition, using a segmentation mask for semantic discovery would fail to find appearance-related attributes.
\citet{zhu2021lowrankgan} propose low-rank subspaces in GANs for image local editing, but the discovery of these subspaces relies on low-rank factorization with a relaxation factor. Such a hyper-parameter turns out to be sensitive to the model structure and the selected local region, and an inadequate value may lead to unsatisfying manipulation results.

Different from existing methods, our algorithm has the following \textbf{\textit{advantages:}}
(1) Our method is based on derivative~\cite{ramesh2018spectral, Chiu2020Jacobian, Wang2021Jacobian} and has no requirements on the model structure as long as it is differentiable.
Hence, unlike some approaches that are particularly designed for StyleGAN~\cite{wu2020stylespace, ling2021editgan}, \method can be easily generalized to different GAN variants.
(2) Our method can be directly solved by maximizing a properly defined generalized Rayleigh quotient, making it independent of any annotations, hyper-parameters, or training.
Such a robust formulation also enables fast implementation, significantly outperforming other alternatives.
(3) Our approach enables more precise local control, which will be verified in the experiment section.

\section{Methodology}\label{sec:method}

In this section, we introduce our proposed method for region-based semantic factorization in GANs.
As mentioned above, we revisit the task of local editing with pre-trained GANs and takes the manipulation model into account for identifying the variation factors regarding a particular image region.
Base on our analysis, the semantic discovery process can be formulated as an optimization problem, whose objective happens to be a well-defined generalized Rayleigh quotient.
Consequently, such a problem holds a super efficient solver.

\subsection{Manipulation Model with GAN Priors}\label{subsec:manipulaton-model}
Using prior knowledge learned by GANs for image editing has been widely explored~\cite{shen2020interfacegan, gansteerability, yang2021semantic}.
Concretely, given a well-trained generator $G(\cdot)$ that maps the latent space $\Z$ to the image space $\X$, we would like to find a latent direction $\n\in\Z$ such that altering a latent code $\z$ through the direction can cause the corresponding semantic change in the output image $\x=G(\z)$.
Such a process can be formulated as
\begin{align}
    \texttt{edit}(\x) \triangleq \x^{\text{edit}} = G(\z + \alpha \n),  \label{eq:manipulation}
\end{align}
where $\alpha$ indicates the degree of editing.
Here, $\n$ is usually assumed to be a unit vector~\cite{shen2021closed}, \textit{i.e.}, $\n^T\n = 1$.

\subsection{Region-based Semantic Discovery}\label{subsec:semantic-discovery}

As shown in \Cref{eq:manipulation}, there is no explicit constraint on the relationship between $\x$ and $\x^{\text{edit}}$, hence the entire image may get changed, resulting in a global editing.
However, for the case of local editing, we would like to only change the image content within a certain region, denoted as $\x_f$, while the surroundings, denoted as $\x_b$, keep untouched.
Here, $\x_f$ and $\x_b$ form a partition of all pixels within $\x$, \textit{i.e.}, $\x_f \cup \x_b = \x$ and $\x_f \cap \x_b = \varnothing$, where the subscripts $f$ and $b$ are short for ``foreground'' and ``background'' respectively.
Accordingly, we reformulate \Cref{eq:manipulation} to fit the local editing task as
\begin{align} \label{eq:local-manipulation}
\begin{cases}
    \x^{\text{edit}} = G(\z + \alpha \n), \\
    \text{s.t.}~||\x_b^{\text{edit}} - \x_b||_2^2 = 0,
\end{cases}
\end{align}
where $||\cdot||_2$ denotes the $\ell_2$ norm.

Based on the local manipulation model described in \Cref{eq:local-manipulation}, the constraint $||\x_b^{\text{edit}} - \x_b||_2^2$ can be approximated using the first-order Taylor expansion
\begin{equation} \label{eq:background}
\begin{aligned}
    ||\x_b^{\text{edit}} - \x_b||_2^2 &= ||\{G(\z + \alpha \n)\}_b - \{G(\z)\}_b||_2^2 \\
                                      &\approx \alpha^2 \n^T \J_b^T\J_b \n.
\end{aligned}
\end{equation}
Here, $\J_b$ is the derivative of pixel values with respect to the latent code (\textit{i.e.}, Jacobian)~\cite{ramesh2018spectral, zhu2021lowrankgan}.
Particularly, we have
\begin{align} \label{eq:jacobian}
    (\J_b)_{j,k} = \frac{ \partial \{G(\z)\}_j }{ \partial \z_k },
\end{align}
where $j$ stands for a pixel position within the image, and $k$ stands for a dimension of the latent space $\Z$.
In practice, $\J_b$ can be easily computed as long as the generator $G(\cdot)$ is differentiable, regardless of its architecture.

In other words, \Cref{eq:background} can be used to characterize the variation of the background region when altering the latent code $\z$ towards the latent direction $\n$.
Similarly, we can also measure the foreground change with
\begin{equation} \label{eq:foreground}
\begin{aligned}
    ||\x_f^{\text{edit}} - \x_f||_2^2 \approx \alpha^2 \n^T \J_f^T\J_f \n.
\end{aligned}
\end{equation}

We argue that, given any arbitrary pixel partition $\{\x_f, \x_b\}$, an adequate local editing should take sufficient effect on the pixels within the region of interest~\cite{shen2021closed}, $\x_f$, yet maintain the remaining pixel values, $\x_b$.
Therefore, we are able to factorize the region-based semantics via solving the following optimization problem
\begin{align} \label{eq:formulation}
\begin{cases}
    \argmax_{\n} \n^T \J_f^T\J_f \n, \\
    \argmin_{\n} ~\n^T \J_b^T\J_b \n. \\
    % \qquad\quad \text{s.t.}~\n^T \n = 1.
\end{cases}
\end{align}

%% Face Results
\begin{figure*}[t]
\centering
\includegraphics[width=0.95\textwidth]{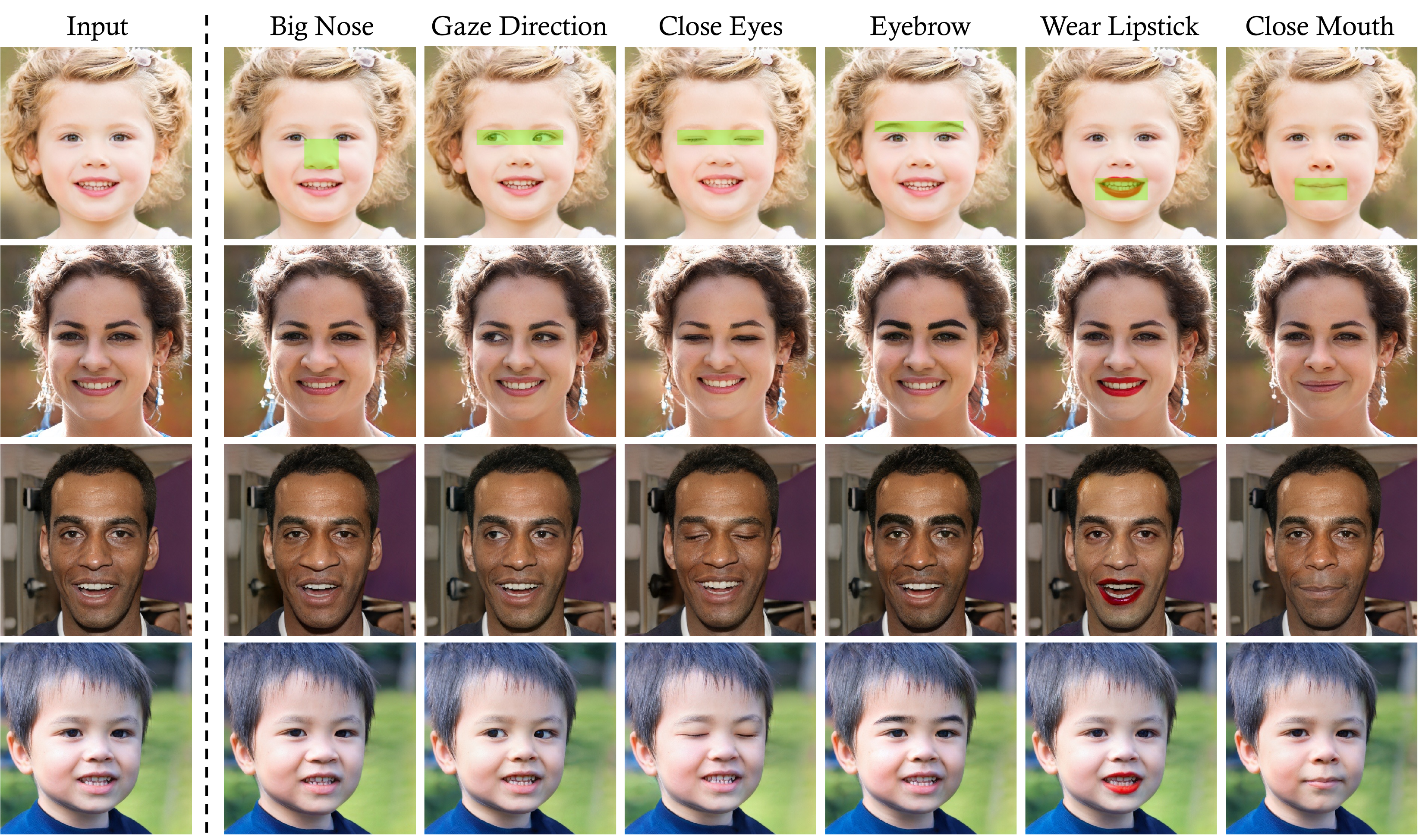}
\vspace{-10pt}
\caption{
    \textbf{Precise local editing results} achieved by our \method on the StyleGAN2 generator~\cite{stylegan2} trained on FFHQ dataset~\cite{stylegan}.
    The regions of interest are highlighted with \textbf{\textcolor{green}{green}} boxes in the first row, while all rows share the same latent directions found by solving \Cref{eq:inverse-equation}.
    Note that our algorithm does \textit{not} require the region masks to be precise, and can identify diverse semantics (\textit{both appearance and shape}) corresponding the same region, like ``wearing lipstick'' and ``closing mouth'' for mouth.
}
\label{fig:face}
\vspace{-5pt}
\end{figure*}

\subsection{Computational Solution}\label{subsec:computational-solution}

\textbf{Reformulation.}
The problem defined in \Cref{eq:formulation} can be solved via convex optimization~\cite{Boyd2004convex}.
Alternatively, we unify this dual-objective optimization into a single criterion as
\begin{align} \label{eq:rayleigh}
    \argmax_{\n} \frac{\n^T \J_f^T\J_f \n}{\n^T \J_b^T\J_b \n},
\end{align}
where the new object happens to be a generalized Rayleigh quotient~\cite{Horn2012matrix}, $R(\J_f, \J_b, \n)$.
\Cref{eq:rayleigh} can be solved as a generalized eigenvalue problem
\begin{align} \label{eq:generalized-eigen}
   \J_f^T\J_f \n = \lambda  \J_b^T\J_b \n,
\end{align}
where those eigenvectors $\n$ corresponding to the largest eigenvalues $\lambda$ are just the local semantics we expect.

\textbf{Standard solution.}
To solve \Cref{eq:generalized-eigen}, a standard solution is to perform Cholesky decomposition~\cite{Higham2002numerical} on $\J_b^T\J_b$ as $\J_b^T\J_b = \mL \mL^T$.
$\mL$ is a lower triangular matrix with real diagonal entries due to the symmetry of $\J_b^T\J_b$.
Let $\tilde{\n} = \mL^T\n$.
\Cref{eq:generalized-eigen} can be reorganized as
\begin{align} 
   \mL^{-1} \J_f^T\J_f (\mL^{-1})^T \tilde{\n} = \lambda \tilde{\n},
\end{align}
which can be easily solved by performing eigen decomposition on $\mL^{-1} \J_f^T\J_f (\mL^{-1})^T$.

\textbf{Handling singular case.}
However, in real cases, there is no guarantee that $\mL$ is invertible.
Recall that the pixel partition $\{\x_f, \x_b\}$ can be arbitrary, making it possible that the variation factors regarding region $\x_b$ are limited.
In other words, $\J_b^T\J_b$ can be rank-deficient.
To handle such a case, we make a slight modification on $\J_b^T\J_b$ to make it non-singular as
\begin{align} 
   \J_b^T\J_b \leftarrow \J_b^T\J_b + \tau \text{tr}(\J_b^T\J_b) \I,
\end{align}
where $\I$ is the identity matrix and $\text{tr}(\cdot)$ denotes the trace.
$\tau=1e^{-3}$ is a small scaling factor.
For simplicity, we set $a = \tau \text{tr}(\J_b^T\J_b)$.
Now, \Cref{eq:generalized-eigen} can be converted to
\begin{align} \label{eq:inverse-equation}
   (\J_b^T\J_b + a\I)^{-1} \J_f^T\J_f \n = \lambda \n.
\end{align}

\textbf{Fast implementation.}
Even though \Cref{eq:inverse-equation} has provided an elegant formulation for region-based semantic factorization, solving it can be time consuming especially when the latent space $\Z$ is with extremely high dimensions.
For example, the $\mathcal{W}^+$ space~\cite{abdal2020image2stylegan++} for a StyleGAN~\cite{stylegan} generator with 18 layers is $512\times18=9216$ dimensional.
To speed up the factorization process, we provide an efficient scheme by virtue of the Sherman-Morrison-Woodbury formula~\cite{Higham2002numerical}.
To be specific, let $\J_b = \U\D\V^T$ be the truncated Singular Value Decomposition (SVD)~\cite{Horn2012matrix} of $\J_b$, where $\U$, $\V$, $\D$ are the left singular matrix, the right singular matrix, and a diagonal matrix respectively.
$\D$ is of size $r\times r$, where $r$ is the rank of $\J_b$.
Then we can derive
\begin{equation} \label{eq:fast-inversion}
\begin{aligned}
      &\ (\J_b^T\J_b + a\I)^{-1} = (\V \D (\V\D)^T + a\I)^{-1} \\
    = &\ a\I - \V\D( a\I + (\V\D)^T \V\D  )^{-1} (\V\D)^T \\
    = &\ a\I - \V\D( a\I + \D^2  )^{-1} \D\V^T \\
    = &\ a\I - \V \tilde{\D}\V^T,
\end{aligned}
\end{equation}
where $\tilde{\D} = \D( a\I + \D^2  )^{-1} \D$ is reduced to the diagonal-wise operation, which is far more efficient.

\textbf{Summary.}
With the above analysis, given a generator $G(\cdot)$ and a partition $\{\x_f, \x_b\}$, the semantic vectors related to region $\x_f$ can be efficiently obtained by solving \Cref{eq:inverse-equation} using \Cref{eq:fast-inversion} as an intermediate step.

\section{Experiments}\label{sec:experiments}

%% Scene car Results
\begin{figure*}[!ht]
\centering
\includegraphics[width=0.95\textwidth]{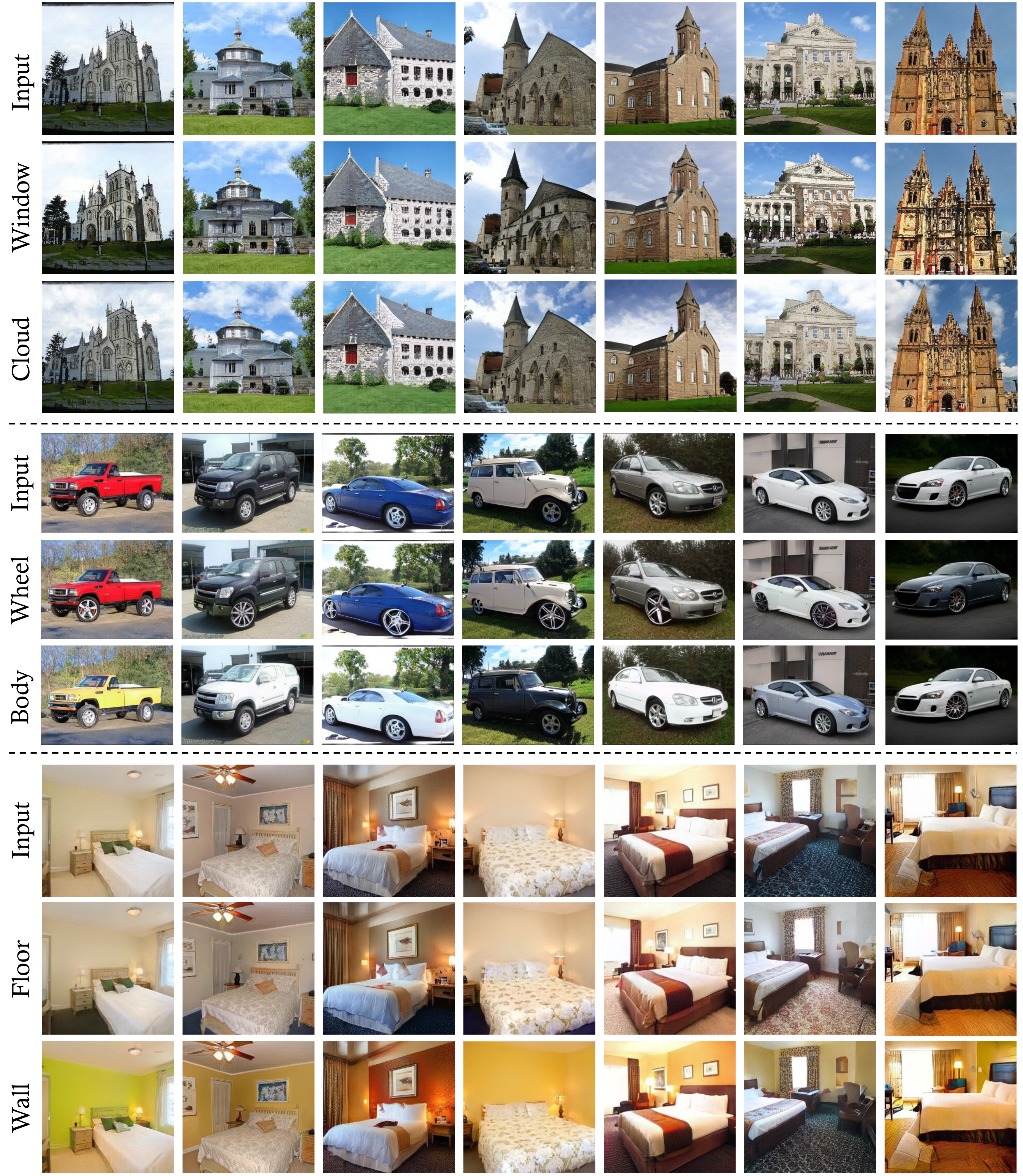}
\vspace{-10pt}
\caption{
    \textbf{Versatile local semantics} found by our algorithm using the StyleGAN2 models~\cite{stylegan2} trained on various datasets, including LSUN churches (indoor scene), LSUN cars (general object), and LSUN bedrooms (indoor scene)~\cite{yu2015lsun}.
}
\label{fig:scene-object}
\vspace{-5pt}
\end{figure*}

%% Bird class on BigGAN 
\begin{figure*}[t]
\centering
\includegraphics[width=0.95\textwidth]{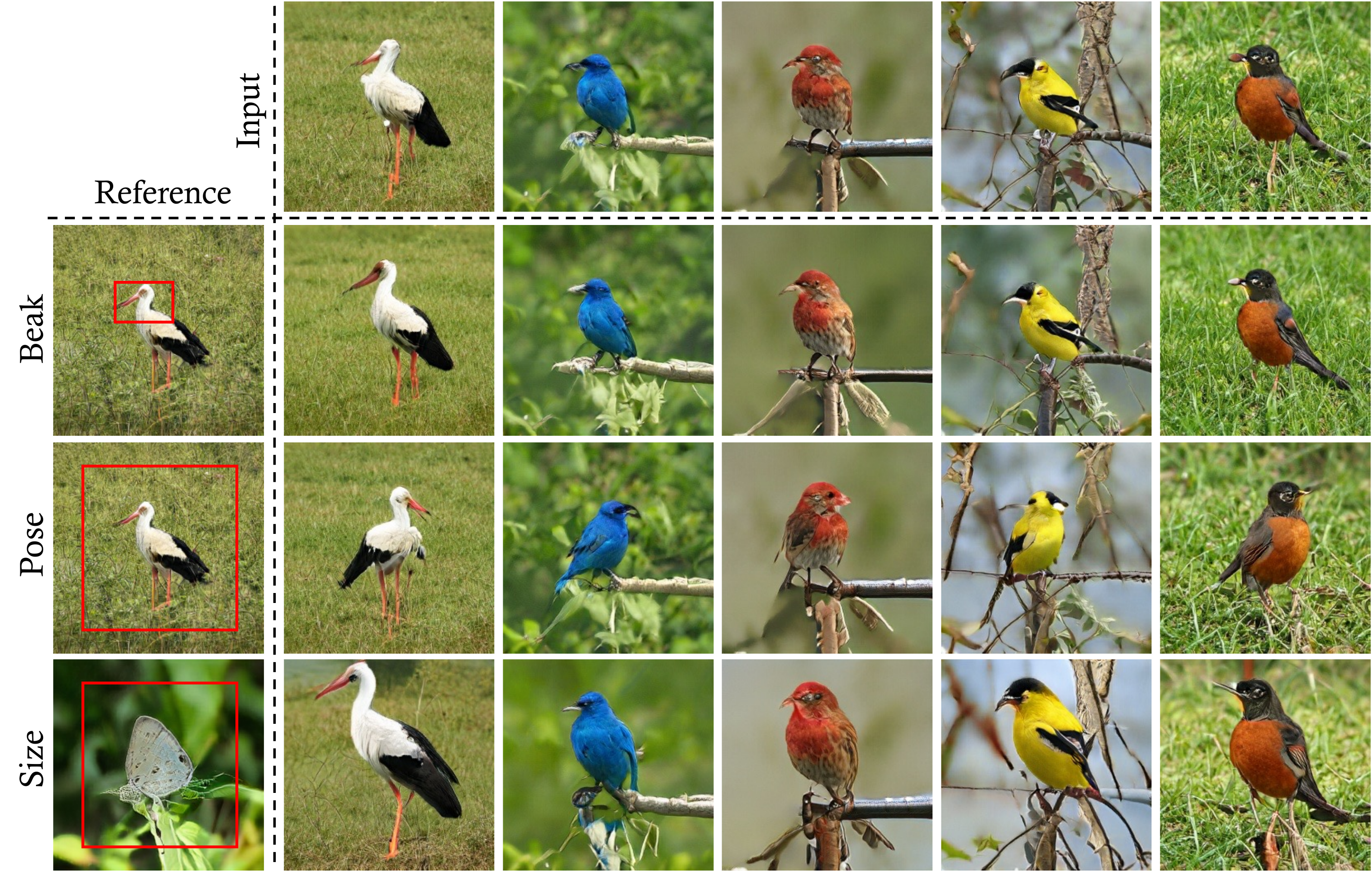}
\vspace{-10pt}
\caption{
    \textbf{Precise local editing on a conditional generative model}, \textit{i.e.}, BigGAN~\cite{biggan}, where we can conclude three observations.
    First, our algorithm is able to control the synthesis of only a part of the object with a small region of interest (\textit{e.g.}, ``beak'' in the second row), or control the synthesis of the entire object with a large region of interest (\textit{e.g.}, ``pose'' in the third row).
    Second, when altering the pose of birds in the third row, the background (\textit{e.g.}, the grass in the second column and the branch in the third column) are barely affected, demonstrating the \textit{precise control} achieved by our method.
    Third, the semantics found from one category can be convincingly applied to other categories.
}
\label{fig:biggan-local-global}
\vspace{-5pt}
\end{figure*}

\subsection{Experimental Setup}
We conduct extensive experiments to evaluate our proposed method, mainly on two types of models, \textit{i.e.}, StyleGAN2~\cite{stylegan2} and BigGAN~\cite{biggan}.
And the datasets we use are diverse, including FFHQ~\cite{stylegan}, LSUN bedroom, church, car~\cite{yu2015lsun}, and ImageNet~\cite{imagenet}.
For StyleGAN2, we use the models released by the authors, and for BigGAN, we use the model from \href{https://www.tensorflow.org/hub/tutorials/biggan\_generation\_with\_tf\_hub}{TensorFlow Hub}.
For metrics, we use Fr\'{e}chet Inception Distance (FID)~\cite{fid}, masked Mean Squared Error (MSE), and Identity loss (ID).
FID is used to evaluate the image fidelity after editing.
For the masked MSE, we used it to qualify the editing precision.
Specifically, we can evaluate the change in the edited region or the rest region using a mask.
We use the \href{https://github.com/TreB1eN/InsightFac\_Pytorch}{ArcFace model} to evaluate identity similarity between the edited images and the original images.
The experiments are organized as follows.
First, we demonstrate that our method could easily find semantically meaningful directions when given a specific region of a generated image on various datasets in \Cref{subsec:precise-editing}.
Second, we compare our method with the existing methods in \Cref{subsec:compare-with-sota} and demonstrate that our methods have strong control over the local region of the synthesized images even when editing in the latent space.
All the experiments are conducted on a single RTX 2080 Ti GPU.

%% Compare with SOTA
\begin{figure*}[t]
\centering
\includegraphics[width=0.95\textwidth]{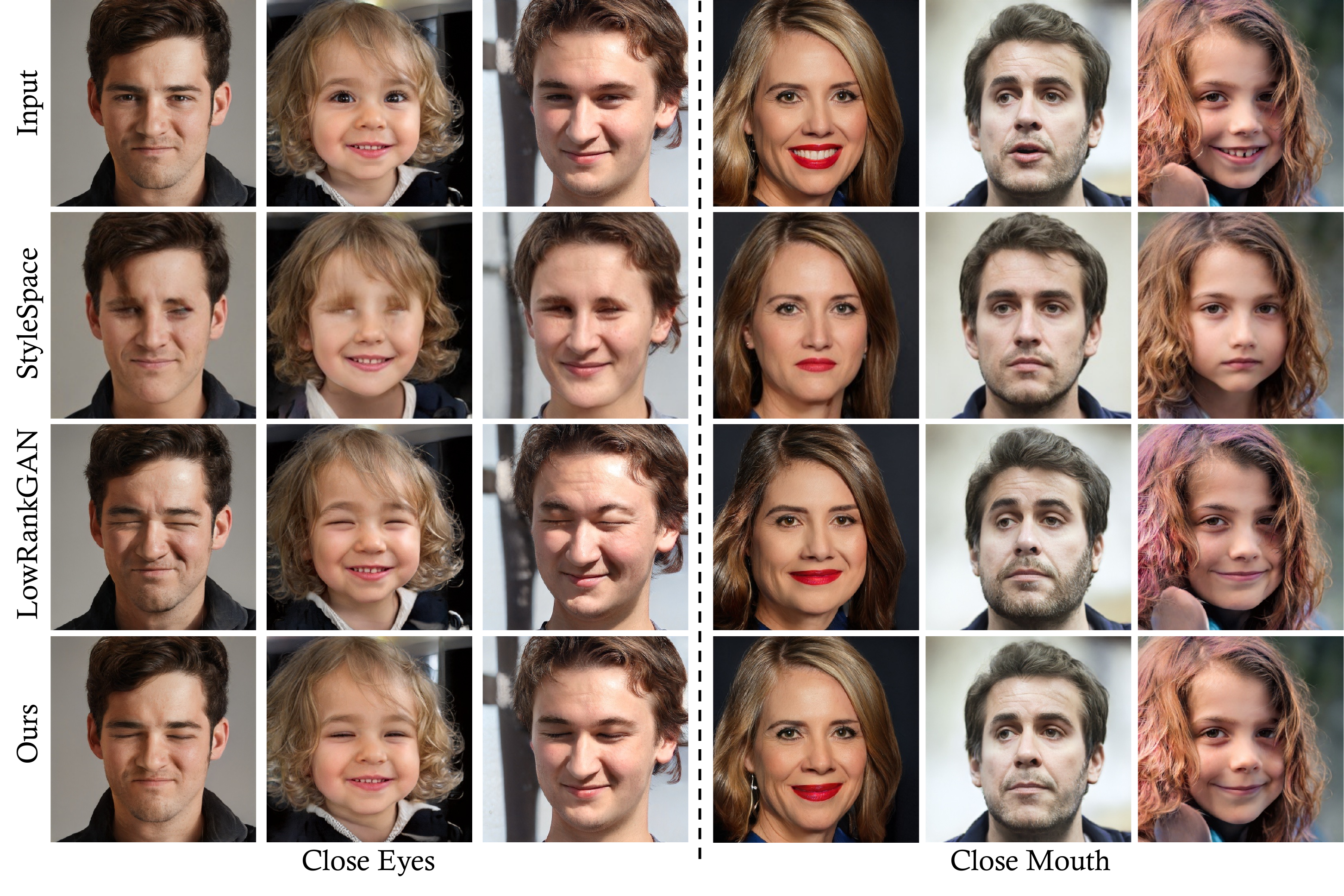}
\vspace{-10pt}
\caption{
    \textbf{Qualitative comparison} between different local editing approaches on face synthesis model~\cite{stylegan2}, including StyleSpace~\cite{wu2020stylespace}, LowRankGAN~\cite{zhu2021lowrankgan}, and \method.
    Our approach produces more realistic results and suggests more precise controllability in maintaining the image contents beyond the region of interests, \textit{i.e.}, eyes and mouth.
}
\label{fig:compare-close-eyes-mouth}
\vspace{-5pt}
\end{figure*}

%%%% Table: Quantitative Comparison on Interpolation and Manipulation
\setlength{\tabcolsep}{6.0pt}
\begin{table*}[t]
  \caption{
    \textbf{Quantitative comparison} between different local editing approaches on face synthesis model~\cite{stylegan2}, including StyleSpace~\cite{wu2020stylespace}, LowRankGAN~\cite{zhu2021lowrankgan}, and \method.
    We use FID (lower is better) to evaluate the image quality after editing, MSE (lower is better, scaled by $1e^4$ for good readability) to calculate the value change of pixels outside the region of interest, and ID similarity (higher is better) to measure the identity change before and after manipulation.
    The interpretation speed using each method is also reported in the last column.
  }
  \vspace{5pt}
  \label{tab:editing-comparison}
  \centering\small
  \begin{tabular}{l|ccc|ccc|ccc|ccc|c}
             \toprule
             &   \multicolumn{3}{c|}{Close Eyes}
             &   \multicolumn{3}{c|}{Close Mouth}
             &   \multicolumn{3}{c|}{With Lipstick}
             &   \multicolumn{3}{c|}{Big Nose}
             &           \\
             \midrule
 Method      &     \FID   &    \MSE    &     \ID     &    \FID   &    \MSE     &    \ID  
             &     \FID   &    \MSE    &     \ID     &    \FID   &    \MSE     &    \ID       &   \textbf{Speed} \\
             \midrule
 StyleSpace 
             &     26.32  &     \textbf{2.31}   &     0.51    &    24.83   &   2.43     &    0.51   
             &     57.12  &     \textbf{0.63}   &     0.84    &    25.65   &  \textbf{ 0.98}     &    0.75      &   10.0s \\
 LowRankGAN
             &    25.43   &     5.61   &    0.53     &    24.91   &    4.96    &    0.73 
             &    \textbf{32.33}   &     8.44   &    0.63     &    25.37   &    5.52    &    0.56      &   393s \\
 \method (Ours)
             &    \textbf{24.40}   &    2.51    &    \textbf{0.83}     &    \textbf{23.35}   &    \textbf{2.11}    &   \textbf{0.85}
             &    38.41   &    1.53    &    \textbf{0.89}     &    \textbf{24.82}   &    1.64    &   \textbf{0.85}       &  0.5s \\
             \bottomrule
  \end{tabular}
  \vspace{0pt}
\end{table*}

\subsection{Local Semantic Discovery and Manipulation}\label{subsec:precise-editing}
Recall that our method is rather simple and can be divided into three steps. 
First, we need to compute the Jacobian of a synthesized image to the latent code.
Second, obtaining $ \J_f $ and $ \J_b $ according to the masked region (By default, in each figure, the masked region and the remaining part are used to compute $ \J_f $ and $ \J_b $, respectively.).
Third, solving \Cref{eq:generalized-eigen} or \Cref{eq:inverse-equation} to get the attribute vectors, which are used to edit the images.
We first conduct experiments on the official FFHQ $ 1024 \times 1024 $ model released in StyleGAN2. 
The eyes, eyebrows, mouth, and nose are chosen as the local regions to factorize their semantics.
As shown in \Cref{fig:face}, our method could uncover the semantics that enables fine-grained and precise local control over the synthesized images.
For example, when factorizing the semantics in the nose-related region, we could find an attribute that changes the size of the nose.
When the region is taken from the eyes, as the green masks are shown in the third and fourth columns in \Cref{fig:face}, several eye-related semantics could be found,  such as look askance, close eyes, etc.
We could manipulate the eyebrows, as the fifth column shows when the region is chosen as the eyebrows. 
The last two columns show the found semantics when given the mouth region. 
For instance, the penultimate column shows that all the images have lipstick regarding their gender.
The last column shows that semantics found in this region can close the mouth of the images. 
More semantic meaningful edit results on each face region can be found in \Cref{appendix:additional-results}.

Besides the face model, we further validate our proposed method on other models.
\Cref{fig:scene-object} shows the results on LSUN-church, car, and bedroom, which demonstrates that our method could control either a large region of an image or a small area of an image.
For the control on the small region, we could observe that the number of windows on the church are modified, and the wheel type of the car are changed as well.
For the control on the large region, the cloud is added in the sky of the church, the color of the car body can be changed, and the color of the wall in bedroom is changed as well.
In all, we could say that our proposed method perform well on StyleGAN2.
For more results, please refer to the \Cref{appendix:additional-results}.

For BigGAN, the most commonly changed attributes are the size, pose, and background color of the objects~\cite{plumerault2020controlling, gansteerability, voynov2020unsupervised}, which always result in a global change of an image. 
Seldom had they shown the local control over the synthesized images.
However, our proposed method not only could control the size or pose of the object, but also could control over a small area of the synthetic images.
\Cref{fig:biggan-local-global} displays the strong ability of our method to edit the local region.
For example, the beak of these birds changed,  which is a very small part of an image.
Still, our method could edit them, which shows the great power of our method for local controllability.
Except for the control on the small region, \Cref{fig:biggan-local-global} also demonstrates that our method could edit the attributes related to a large region as well.
For instance, the size and pose can be varied successfully regrading the categories.

\subsection{Comparison with Existing Alternatives} \label{subsec:compare-with-sota}
In this section, we compare our method with the state-of-the-art methods both qualitatively and quantitatively.
We compare our method with StyleSpace~\cite{wu2020stylespace} and LowRankGAN~\cite{zhu2021lowrankgan}, which are two state-of-the-art methods for local control on the generated images.
In the main paper, we show the attributes of closing eyes and mouth, and for the comparison of the other attribute, please see the  the \Cref{appendix:additional-results}.

First, we compare with StyleSpace as shown in \Cref{fig:compare-close-eyes-mouth}, in which we could find that our method could achieve more photo-realistic results.
For instance, the artifacts appeared when closing the eyes of the man, and the eyebrow disappeared simultaneously.
The hair and background of the second image are changing as well.
Sometimes, it is hard to close the images' eyes, as the second column shows.
When closing the mouth, the jaw becomes smaller in all these three images, and the beard of the man decreases as well.
On the contrary, the editing results are significantly improved when using our method. 

Second, we compared our method with recently proposed LowRankGAN~\cite{zhu2021lowrankgan}.
As shown in \Cref{fig:compare-close-eyes-mouth}, we could see that both the LowRankGAN and our method could successfully close the eyes or mouth  of the images.
Nevertheless, there are some differences.
For one thing, when closing the eyes of the first and third images, the brightness changes, the face of the second image is smaller after editing.
Instead, our method could well preserve these changes.
For another, when closing the mouth, the jaw of the first woman is widened, the beard of the man is increased, and the hair color and the background of the girl are changed. 
Again, our method could well preserve these changes.
We draw from the above experiment that our method has a stronger ability to precisely manipulate the local regions than those two baselines.

\begin{figure}[t]
\centering
\includegraphics[width=1.0\linewidth]{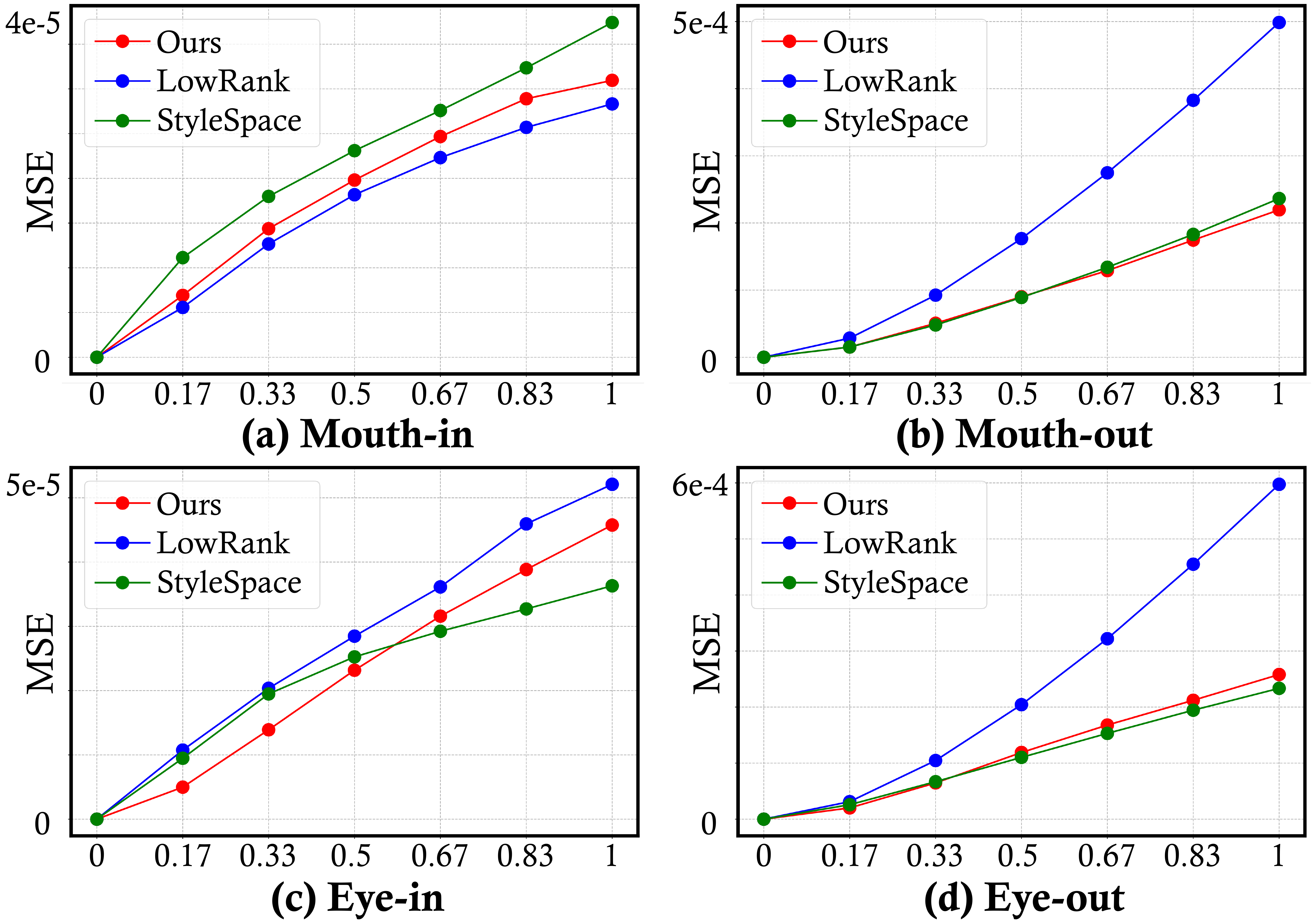}
\vspace{-20pt}
\caption{
    \textbf{Quantitative results of pixel change} by \textit{gradually} making people close mouth and eyes using StyleSpace~\cite{wu2020stylespace}, LowRankGAN~\cite{zhu2021lowrankgan}, and \method.
    ``in'' and ``out'' refer to region of interest and its surroundings, respectively.
    A higher ``in'' change and a lower ``out'' change are expected for a promising local editing.
}
\label{fig:compare-close-eyes-mouth-curve}
\vspace{-5pt}
\end{figure}

We also give the quantitative comparison results in \Cref{tab:editing-comparison} and \Cref{fig:compare-close-eyes-mouth-curve}.
\Cref{tab:editing-comparison} reports quantitative results on different attributes.
As shown, our method could get the best identity similarity (ID) after editing for all these attributes.
Regrading the speed to find the semantics in a specific region, we also report the time of discovery using different methods in \Cref{tab:editing-comparison} (The time ).
It can be observed that our \method owns the fastest implementation, thanks to the analysis on the local manipulation model.
Also, it is worth noting that our algorithm does not require any annotations like StyleSpace~\cite{wu2020stylespace}.

\Cref{fig:compare-close-eyes-mouth-curve} gives the MSE on both the edited region and the remaining region when gradually increasing the manipulation strength (\textit{i.e.}, $ \alpha $ in \Cref{eq:manipulation}) on closing mouth or eyes.
\Cref{fig:compare-close-eyes-mouth-curve}a shows that the MSE of StyleSpace is larger than the other two methods in the mouth region, presumably because the jaw has a large shift when closing the mouth, as shown in \Cref{fig:compare-close-eyes-mouth}.
\Cref{fig:compare-close-eyes-mouth-curve}b shows that the MSE of LowRankGAN is far better than the other two methods.
When it comes to \Cref{fig:compare-close-eyes-mouth-curve}c and \Cref{fig:compare-close-eyes-mouth-curve}d, the MSE of StyleSpace in the eye region is small, while LowRankGAN gets the largest MSE again.
Hence, our method could better control the specific region since the change in the region is large while the change in the remaining region is small compared to the state-of-the-art methods.

%% Masked curve with SOTA

\subsection{Discussion}
We have demonstrated the impressive ability of our method in local control, but there are still some limitations.
For example, our approach would fail to control extremely small regions embedded in large area of holistically uniform textures (\textit{e.g.}, one tooth), or regions with multiple similar components in the image (\textit{e.g.}, a single nostril) very well.
It also shares a common limitation as existing methods in editing only one object of a symmetric pair (\textit{e.g.}, one eye of a human face).
Meanwhile, it is hard to discover the global semantic directions in StyleGAN (\textit{e.g.}, face pose), even choosing a sufficiently large region.
Future research will focus on how to generalize our region-based semantic factorization on more fine-grained regions as well as how to unify global and local semantic exploration within the same algorithm.

\section{Conclusion}\label{sec:conclusion}

In this work, we propose a simple algorithm to factorize the semantics learned by GANs regarding some particular regions.
We re-examine the task of local editing and take the manipulation model into account for semantic discovery. 
By appropriately formulating this process as a dual-objective optimization problem, we enable an efficient and robust algorithm to find region-based variation factors without relying on any annotations or training.
%

%%%% References
\bibliography{ref}

\begin{thebibliography}{52}
\providecommand{\natexlab}[1]{#1}
\providecommand{\url}[1]{\texttt{#1}}
\expandafter\ifx\csname urlstyle\endcsname\relax
  \providecommand{\doi}[1]{doi: #1}\else
  \providecommand{\doi}{doi: \begingroup \urlstyle{rm}\Url}\fi

\bibitem[Abdal et~al.(2020)Abdal, Qin, and Wonka]{abdal2020image2stylegan++}
Abdal, R., Qin, Y., and Wonka, P.
\newblock Image2stylegan++: How to edit the embedded images?
\newblock In \emph{IEEE Conf. Comput. Vis. Pattern Recog.}, 2020.

\bibitem[Arjovsky et~al.(2017)Arjovsky, Chintala, and Bottou]{wgan}
Arjovsky, M., Chintala, S., and Bottou, L.
\newblock Wasserstein generative adversarial networks.
\newblock In \emph{ICML}, 2017.

\bibitem[Bau et~al.(2019)Bau, Zhu, Strobelt, Zhou, Tenenbaum, Freeman, and
  Torralba]{bau2019gandissection}
Bau, D., Zhu, J.-Y., Strobelt, H., Zhou, B., Tenenbaum, J.~B., Freeman, W.~T.,
  and Torralba, A.
\newblock {GAN} dissection: Visualizing and understanding generative
  adversarial networks.
\newblock In \emph{Int. Conf. Learn. Represent.}, 2019.

\bibitem[Bau et~al.(2020{\natexlab{a}})Bau, Liu, Wang, Zhu, and
  Torralba]{bau2020rewriting}
Bau, D., Liu, S., Wang, T., Zhu, J.-Y., and Torralba, A.
\newblock Rewriting a deep generative model.
\newblock In \emph{Eur. Conf. Comput. Vis.}, 2020{\natexlab{a}}.

\bibitem[Bau et~al.(2020{\natexlab{b}})Bau, Strobelt, Peebles, Wulff, Zhou,
  Zhu, and Torralba]{bau2020ganpaint}
Bau, D., Strobelt, H., Peebles, W., Wulff, J., Zhou, B., Zhu, J.-Y., and
  Torralba, A.
\newblock Semantic photo manipulation with a generative image prior.
\newblock \emph{ACM Trans. Graph.}, 2020{\natexlab{b}}.

\bibitem[Boyd \& Vandenberghe(2004)Boyd and Vandenberghe]{Boyd2004convex}
Boyd, S. and Vandenberghe, L.
\newblock \emph{Convex Optimization}.
\newblock Cambridge University Press, 2004.

\bibitem[Brock et~al.(2019)Brock, Donahue, and Simonyan]{biggan}
Brock, A., Donahue, J., and Simonyan, K.
\newblock Large scale {GAN} training for high fidelity natural image synthesis.
\newblock In \emph{Int. Conf. Learn. Represent.}, 2019.

\bibitem[Cherepkov et~al.(2021)Cherepkov, Voynov, and
  Babenko]{cherepkov2021navigate}
Cherepkov, A., Voynov, A., and Babenko, A.
\newblock Navigating the {GAN} parameter space for semantic image editing.
\newblock In \emph{IEEE Conf. Comput. Vis. Pattern Recog.}, 2021.

\bibitem[Chiu et~al.(2020)Chiu, Koyama, Lai, Igarashi, and
  Yue]{Chiu2020Jacobian}
Chiu, C.-H., Koyama, Y., Lai, Y., Igarashi, T., and Yue, Y.
\newblock Human-in-the-loop differential subspace search in high-dimensional
  latent space.
\newblock \emph{ACM Trans. Graph.}, 2020.

\bibitem[Collins et~al.(2020)Collins, Bala, Price, and
  Susstrunk]{collins2020editing}
Collins, E., Bala, R., Price, B., and Susstrunk, S.
\newblock Editing in style: Uncovering the local semantics of {GANs}.
\newblock In \emph{IEEE Conf. Comput. Vis. Pattern Recog.}, 2020.

\bibitem[Deng et~al.(2009)Deng, Dong, Socher, Li, Li, and Fei-Fei]{imagenet}
Deng, J., Dong, W., Socher, R., Li, L.-J., Li, K., and Fei-Fei, L.
\newblock {ImageNet}: A large-scale hierarchical image database.
\newblock In \emph{IEEE Conf. Comput. Vis. Pattern Recog.}, 2009.

\bibitem[Ghojogh et~al.(2019)Ghojogh, Karray, and Crowley]{reyleigh-quotient}
Ghojogh, B., Karray, F., and Crowley, M.
\newblock Eigenvalue and generalized eigenvalue problems: Tutorial, 2019.

\bibitem[Goetschalckx et~al.(2019)Goetschalckx, Andonian, Oliva, and
  Isola]{goetschalckx2019ganalyze}
Goetschalckx, L., Andonian, A., Oliva, A., and Isola, P.
\newblock {GANalyze}: Toward visual definitions of cognitive image properties.
\newblock In \emph{Int. Conf. Comput. Vis.}, 2019.

\bibitem[Goodfellow et~al.(2014)Goodfellow, Pouget-Abadie, Mirza, Xu,
  Warde-Farley, Ozair, Courville, and Bengio]{gan}
Goodfellow, I., Pouget-Abadie, J., Mirza, M., Xu, B., Warde-Farley, D., Ozair,
  S., Courville, A., and Bengio, Y.
\newblock Generative adversarial networks.
\newblock In \emph{Adv. Neural Inform. Process. Syst.}, 2014.

\bibitem[Gu et~al.(2020)Gu, Shen, and Zhou]{gu2020image}
Gu, J., Shen, Y., and Zhou, B.
\newblock Image processing using multi-code gan prior.
\newblock In \emph{IEEE Conf. Comput. Vis. Pattern Recog.}, 2020.

\bibitem[H{\"a}rk{\"o}nen et~al.(2020)H{\"a}rk{\"o}nen, Hertzmann, Lehtinen,
  and Paris]{ganspace}
H{\"a}rk{\"o}nen, E., Hertzmann, A., Lehtinen, J., and Paris, S.
\newblock {GANSpace}: Discovering interpretable {GAN} controls.
\newblock In \emph{Adv. Neural Inform. Process. Syst.}, 2020.

\bibitem[He et~al.(2021)He, Kan, and Shan]{he2021eigengan}
He, Z., Kan, M., and Shan, S.
\newblock {EigenGAN}: Layer-wise eigen-learning for {GAN}s.
\newblock In \emph{Int. Conf. Comput. Vis.}, 2021.

\bibitem[Heusel et~al.(2017)Heusel, Ramsauer, Unterthiner, Nessler, and
  Hochreiter]{fid}
Heusel, M., Ramsauer, H., Unterthiner, T., Nessler, B., and Hochreiter, S.
\newblock {GANs} trained by a two time-scale update rule converge to a local
  nash equilibrium.
\newblock In \emph{Adv. Neural Inform. Process. Syst.}, 2017.

\bibitem[Higham(2002)]{Higham2002numerical}
Higham, N.~J.
\newblock \emph{Accuracy and Stability of Numerical Algorithms}.
\newblock SIAM, 2002.

\bibitem[Horn \& Johnson(2012)Horn and Johnson]{Horn2012matrix}
Horn, R.~A. and Johnson, C.~R.
\newblock \emph{Matrix Analysis}.
\newblock Cambridge University Press, 2012.

\bibitem[Jahanian et~al.(2020)Jahanian, Chai, and Isola]{gansteerability}
Jahanian, A., Chai, L., and Isola, P.
\newblock On the" steerability" of generative adversarial networks.
\newblock In \emph{Int. Conf. Learn. Represent.}, 2020.

\bibitem[Karras et~al.(2018)Karras, Aila, Laine, and Lehtinen]{pggan}
Karras, T., Aila, T., Laine, S., and Lehtinen, J.
\newblock Progressive growing of {GAN}s for improved quality, stability, and
  variation.
\newblock In \emph{Int. Conf. Learn. Represent.}, 2018.

\bibitem[Karras et~al.(2019)Karras, Laine, and Aila]{stylegan}
Karras, T., Laine, S., and Aila, T.
\newblock A style-based generator architecture for generative adversarial
  networks.
\newblock In \emph{IEEE Conf. Comput. Vis. Pattern Recog.}, 2019.

\bibitem[Karras et~al.(2020{\natexlab{a}})Karras, Aittala, Hellsten, Laine,
  Lehtinen, and Aila]{stylegan2ada}
Karras, T., Aittala, M., Hellsten, J., Laine, S., Lehtinen, J., and Aila, T.
\newblock Training generative adversarial networks with limited data.
\newblock In \emph{Adv. Neural Inform. Process. Syst.}, 2020{\natexlab{a}}.

\bibitem[Karras et~al.(2020{\natexlab{b}})Karras, Laine, Aittala, Hellsten,
  Lehtinen, and Aila]{stylegan2}
Karras, T., Laine, S., Aittala, M., Hellsten, J., Lehtinen, J., and Aila, T.
\newblock Analyzing and improving the image quality of {StyleGAN}.
\newblock In \emph{IEEE Conf. Comput. Vis. Pattern Recog.}, 2020{\natexlab{b}}.

\bibitem[Karras et~al.(2021)Karras, Aittala, Laine, H\"ark\"onen, Hellsten,
  Lehtinen, and Aila]{stylegan3}
Karras, T., Aittala, M., Laine, S., H\"ark\"onen, E., Hellsten, J., Lehtinen,
  J., and Aila, T.
\newblock Alias-free generative adversarial networks.
\newblock In \emph{Adv. Neural Inform. Process. Syst.}, 2021.

\bibitem[Kim et~al.(2021)Kim, Choi, Kim, Yoo, and Uh]{kim2021exploiting}
Kim, H., Choi, Y., Kim, J., Yoo, S., and Uh, Y.
\newblock Exploiting spatial dimensions of latent in gan for real-time image
  editing.
\newblock In \emph{IEEE Conf. Comput. Vis. Pattern Recog.}, 2021.

\bibitem[Lang et~al.(2021)Lang, Gandelsman, Yarom, Wald, Elidan, Hassidim,
  Freeman, Isola, Globerson, Irani, et~al.]{lang2021explaining}
Lang, O., Gandelsman, Y., Yarom, M., Wald, Y., Elidan, G., Hassidim, A.,
  Freeman, W.~T., Isola, P., Globerson, A., Irani, M., et~al.
\newblock Explaining in style: Training a gan to explain a classifier in
  stylespace.
\newblock In \emph{Int. Conf. Comput. Vis.}, 2021.

\bibitem[Lee et~al.(2020)Lee, Liu, Wu, and Luo]{lee2020maskgan}
Lee, C.-H., Liu, Z., Wu, L., and Luo, P.
\newblock Maskgan: Towards diverse and interactive facial image manipulation.
\newblock In \emph{IEEE Conf. Comput. Vis. Pattern Recog.}, 2020.

\bibitem[Ling et~al.(2021)Ling, Kreis, Li, Kim, Torralba, and
  Fidler]{ling2021editgan}
Ling, H., Kreis, K., Li, D., Kim, S.~W., Torralba, A., and Fidler, S.
\newblock {EditGAN}: High-precision semantic image editing.
\newblock In \emph{Adv. Neural Inform. Process. Syst.}, 2021.

\bibitem[Menon et~al.(2020)Menon, Damian, Hu, Ravi, and Rudin]{menon2020pulse}
Menon, S., Damian, A., Hu, S., Ravi, N., and Rudin, C.
\newblock Pulse: Self-supervised photo upsampling via latent space exploration
  of generative models.
\newblock In \emph{IEEE Conf. Comput. Vis. Pattern Recog.}, 2020.

\bibitem[Peebles et~al.(2021)Peebles, Zhu, Zhang, Torralba, Efros, and
  Shechtman]{peebles2021gan}
Peebles, W., Zhu, J.-Y., Zhang, R., Torralba, A., Efros, A., and Shechtman, E.
\newblock Gan-supervised dense visual alignment.
\newblock \emph{arXiv preprint arXiv:2112.05143}, 2021.

\bibitem[Plumerault et~al.(2020)Plumerault, Borgne, and
  Hudelot]{plumerault2020controlling}
Plumerault, A., Borgne, H.~L., and Hudelot, C.
\newblock Controlling generative models with continuous factors of variations.
\newblock In \emph{Int. Conf. Learn. Represent.}, 2020.

\bibitem[Ramesh et~al.(2019)Ramesh, Choi, and LeCun]{ramesh2018spectral}
Ramesh, A., Choi, Y., and LeCun, Y.
\newblock A spectral regularizer for unsupervised disentanglement.
\newblock In \emph{Int. Conf. Mach. Learn.}, 2019.

\bibitem[Shen \& Zhou(2021)Shen and Zhou]{shen2021closed}
Shen, Y. and Zhou, B.
\newblock Closed-form factorization of latent semantics in {GANs}.
\newblock In \emph{IEEE Conf. Comput. Vis. Pattern Recog.}, 2021.

\bibitem[Shen et~al.(2020{\natexlab{a}})Shen, Gu, Tang, and
  Zhou]{shen2020interpreting}
Shen, Y., Gu, J., Tang, X., and Zhou, B.
\newblock Interpreting the latent space of {GAN}s for semantic face editing.
\newblock In \emph{IEEE Conf. Comput. Vis. Pattern Recog.}, 2020{\natexlab{a}}.

\bibitem[Shen et~al.(2020{\natexlab{b}})Shen, Yang, Tang, and
  Zhou]{shen2020interfacegan}
Shen, Y., Yang, C., Tang, X., and Zhou, B.
\newblock {InterFaceGAN}: Interpreting the disentangled face representation
  learned by {GANs}.
\newblock \emph{IEEE Trans. Pattern Anal. Mach. Intell.}, 2020{\natexlab{b}}.

\bibitem[Spingarn-Eliezer et~al.(2021)Spingarn-Eliezer, Banner, and
  Michaeli]{spingarn2021gan}
Spingarn-Eliezer, N., Banner, R., and Michaeli, T.
\newblock {GAN} steerability without optimization.
\newblock In \emph{Int. Conf. Learn. Represent.}, 2021.

\bibitem[Suzuki et~al.(2018)Suzuki, Koyama, Miyato, Yonetsuji, and
  Zhu]{suzuki2018spatially}
Suzuki, R., Koyama, M., Miyato, T., Yonetsuji, T., and Zhu, H.
\newblock Spatially controllable image synthesis with internal representation
  collaging.
\newblock \emph{arXiv preprint arXiv:1811.10153}, 2018.

\bibitem[Tan et~al.(2020)Tan, Shen, and Zhou]{tan2020improving}
Tan, S., Shen, Y., and Zhou, B.
\newblock Improving the fairness of deep generative models without retraining.
\newblock \emph{arXiv preprint arXiv:2012.04842}, 2020.

\bibitem[Voynov \& Babenko(2020)Voynov and Babenko]{voynov2020unsupervised}
Voynov, A. and Babenko, A.
\newblock Unsupervised discovery of interpretable directions in the {GAN}
  latent space.
\newblock In \emph{Int. Conf. Mach. Learn.}, 2020.

\bibitem[Wang \& Ponce(2021)Wang and Ponce]{Wang2021Jacobian}
Wang, B. and Ponce, C.~R.
\newblock The geometry of deep generative image models and its applications.
\newblock In \emph{Int. Conf. Learn. Represent.}, 2021.

\bibitem[Wu et~al.(2021)Wu, Lischinski, and Shechtman]{wu2020stylespace}
Wu, Z., Lischinski, D., and Shechtman, E.
\newblock {StyleSpace} analysis: Disentangled controls for {StyleGAN} image
  generation.
\newblock In \emph{IEEE Conf. Comput. Vis. Pattern Recog.}, 2021.

\bibitem[Xu et~al.(2021)Xu, Shen, Zhu, Yang, and Zhou]{xu2021generative}
Xu, Y., Shen, Y., Zhu, J., Yang, C., and Zhou, B.
\newblock Generative hierarchical features from synthesizing images.
\newblock In \emph{IEEE Conf. Comput. Vis. Pattern Recog.}, 2021.

\bibitem[Yang et~al.(2021{\natexlab{a}})Yang, Shen, Xu, and Zhou]{yang2021data}
Yang, C., Shen, Y., Xu, Y., and Zhou, B.
\newblock Data-efficient instance generation from instance discrimination.
\newblock In \emph{2021}, 2021{\natexlab{a}}.

\bibitem[Yang et~al.(2021{\natexlab{b}})Yang, Shen, and Zhou]{yang2021semantic}
Yang, C., Shen, Y., and Zhou, B.
\newblock Semantic hierarchy emerges in deep generative representations for
  scene synthesis.
\newblock \emph{Int. J. Comput. Vis.}, 2021{\natexlab{b}}.

\bibitem[Yu et~al.(2015)Yu, Seff, Zhang, Song, Funkhouser, and
  Xiao]{yu2015lsun}
Yu, F., Seff, A., Zhang, Y., Song, S., Funkhouser, T., and Xiao, J.
\newblock {LSUN}: Construction of a large-scale image dataset using deep
  learning with humans in the loop.
\newblock \emph{arXiv preprint arXiv:1506.03365}, 2015.

\bibitem[Zhang et~al.(2021{\natexlab{a}})Zhang, Xu, and
  Shen]{zhang2021decorating}
Zhang, C., Xu, Y., and Shen, Y.
\newblock Decorating your own bedroom: Locally controlling image generation
  with generative adversarial networks.
\newblock \emph{arXiv preprint arXiv:2105.08222}, 2021{\natexlab{a}}.

\bibitem[Zhang et~al.(2021{\natexlab{b}})Zhang, Ling, Gao, Yin, Lafleche,
  Barriuso, Torralba, and Fidler]{zhang2021datasetgan}
Zhang, Y., Ling, H., Gao, J., Yin, K., Lafleche, J.-F., Barriuso, A., Torralba,
  A., and Fidler, S.
\newblock Datasetgan: Efficient labeled data factory with minimal human effort.
\newblock In \emph{IEEE Conf. Comput. Vis. Pattern Recog.}, 2021{\natexlab{b}}.

\bibitem[Zhao et~al.(2020)Zhao, Liu, Lin, Zhu, and Han]{zhao2020differentiable}
Zhao, S., Liu, Z., Lin, J., Zhu, J.-Y., and Han, S.
\newblock Differentiable augmentation for data-efficient gan training.
\newblock In \emph{Adv. Neural Inform. Process. Syst.}, 2020.

\bibitem[Zhu et~al.(2020)Zhu, Shen, Zhao, and Zhou]{zhu2020domain}
Zhu, J., Shen, Y., Zhao, D., and Zhou, B.
\newblock In-domain {GAN} inversion for real image editing.
\newblock \emph{Eur. Conf. Comput. Vis.}, 2020.

\bibitem[Zhu et~al.(2021)Zhu, Feng, Shen, Zhao, Zha, Zhou, and
  Chen]{zhu2021lowrankgan}
Zhu, J., Feng, R., Shen, Y., Zhao, D., Zha, Z., Zhou, J., and Chen, Q.
\newblock Low-rank subspaces in {GAN}s.
\newblock In \emph{Adv. Neural Inform. Process. Syst.}, 2021.

\end{thebibliography}
\bibliographystyle{icml2022}

%%%% Appendix
\newpage
\appendix
\onecolumn
\section*{Appendix}
\renewcommand{\thefigure}{A\arabic{figure}}
\setcounter{figure}{0}

\section{Overview}\label{appendix:overview}
This paper proposed ReSeFa to precisely control the local regions of the synthesized images on pre-trained GANs.
This appendix is organized as follows.
First, we give some proofs of the equations in the main paper in \Cref{appendix:proof}.
Second, we provide more qualitative results in \Cref{appendix:additional-results} to demonstrate the effectiveness of our method.

\section{Proof}\label{appendix:proof}
The proof of \Cref{eq:rayleigh} $\to$ \Cref{eq:generalized-eigen} is listed as follows.

According to the Rayleigh-Ritz quotient method~\cite{reyleigh-quotient}, the optimization problem in \Cref{eq:rayleigh} can be cast as
\begin{equation}
    \begin{cases}
    \argmax_{\n} \n^T \J_f^T\J_f \n, \\
    \qquad\quad \text{s.t.}~~\n^T \J_b^T\J_b \n = 1.
\end{cases}
\end{equation}
Hence, the Lagrangian~\cite{Boyd2004convex} is
\begin{equation} \label{eq:largange}
    \mathcal{L} = \n^T \J_f^T\J_f \n - \lambda (\n^T \J_b^T\J_b \n - 1),
\end{equation}
where $\lambda$ is the Lagrange multiplier. Performing the derivatives of \Cref{eq:largange} on $\n$, we can get
\begin{equation} \label{eq:derivate}
    \frac{\partial \mathcal{L}}{ \partial \n} = 2 \J_f^T\J_f \n - 2 \lambda \J_b^T\J_b \n 
\end{equation}
Setting \Cref{eq:derivate} equals to zero, we have
\begin{equation}
    \J_f^T\J_f \n  =  \lambda \J_b^T\J_b \n .
\end{equation}

The deduction of Eq.~(8) $\to$ Eq.~(9) is outlined as follows.

Substituting $\J_b^T\J_b = \mL \mL^T$ into Eq.~(8), we have
\begin{equation}
    \J_f^T\J_f \n = \lambda  \mL \mL^T \n.
\end{equation}
Multiplying both sides with $\mL^{-1}$ results in
\begin{gather}
    \mL^{-1}\J_f^T\J_f \n = \lambda \mL^T \n.
\end{gather}
We can further write
\begin{gather}
    \mL^{-1}\J_f^T\J_f (\mL^{-1})^T \mL^T \n = \lambda \mL^T \n.
\end{gather}
Letting $\tilde{\n} = \mL^T \n$ delivers Eq.~(9).

\section{Additional Results}\label{appendix:additional-results}
\begin{figure*}[!ht]
\centering
\includegraphics[width=0.95\textwidth]{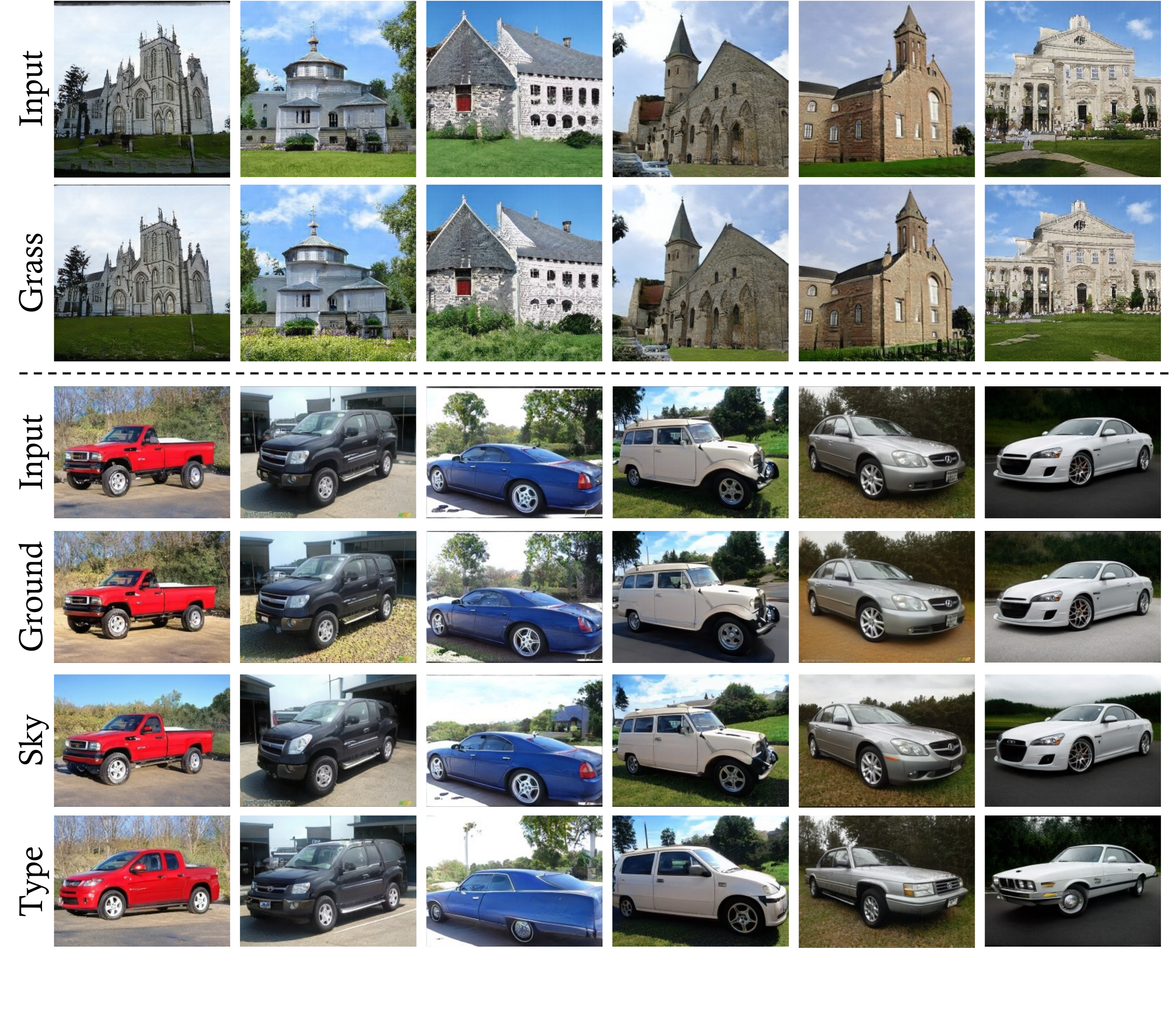}
\vspace{-5pt}
\caption{\textbf{Versatile local semantics} found by our algorithm using the StyleGAN2 models~\cite{stylegan2} trained on various datasets, including LSUN churches (indoor scene) and LSUN cars (general object)~\cite{yu2015lsun}.}
\label{fig-appendix:scene-car}
\vspace{-5pt}
\end{figure*}

\begin{figure*}[!ht]
\centering
\includegraphics[width=0.95\textwidth]{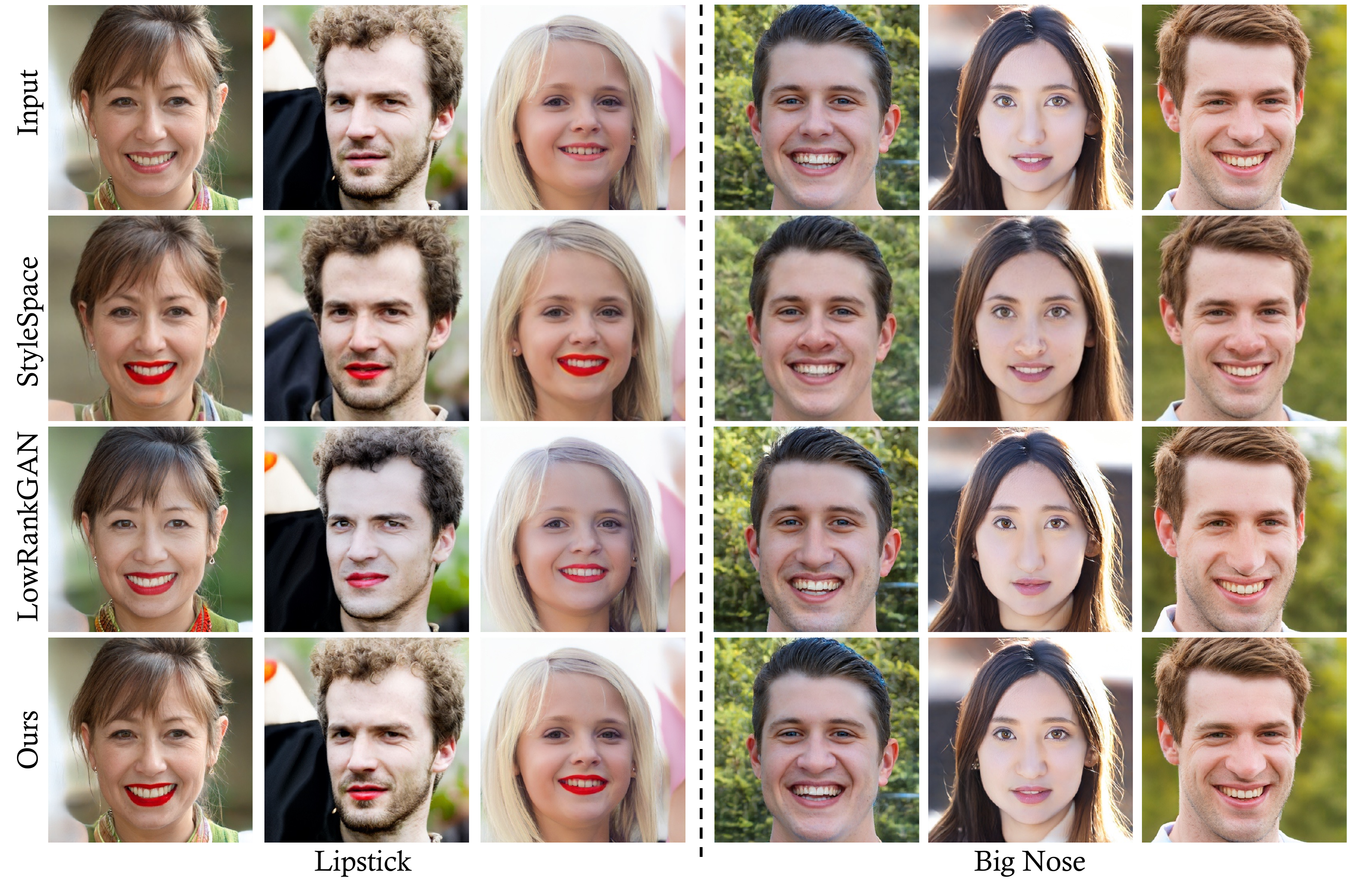}
\vspace{-10pt}
\caption{
    \textbf{Qualitative comparison} between different local editing approaches on face synthesis model~\cite{stylegan2}, including StyleSpace~\cite{wu2020stylespace}, LowRankGAN~\cite{zhu2021lowrankgan}, and \method.
    Our approach produces more realistic results and suggests more precise controllability in maintaining the image contents beyond the region of interests, \textit{i.e.}, mouth and nose.
}
\label{fig-appendix:compare}
\vspace{-5pt}
\end{figure*}

\begin{figure*}[!ht]
\centering
\includegraphics[width=0.95\textwidth]{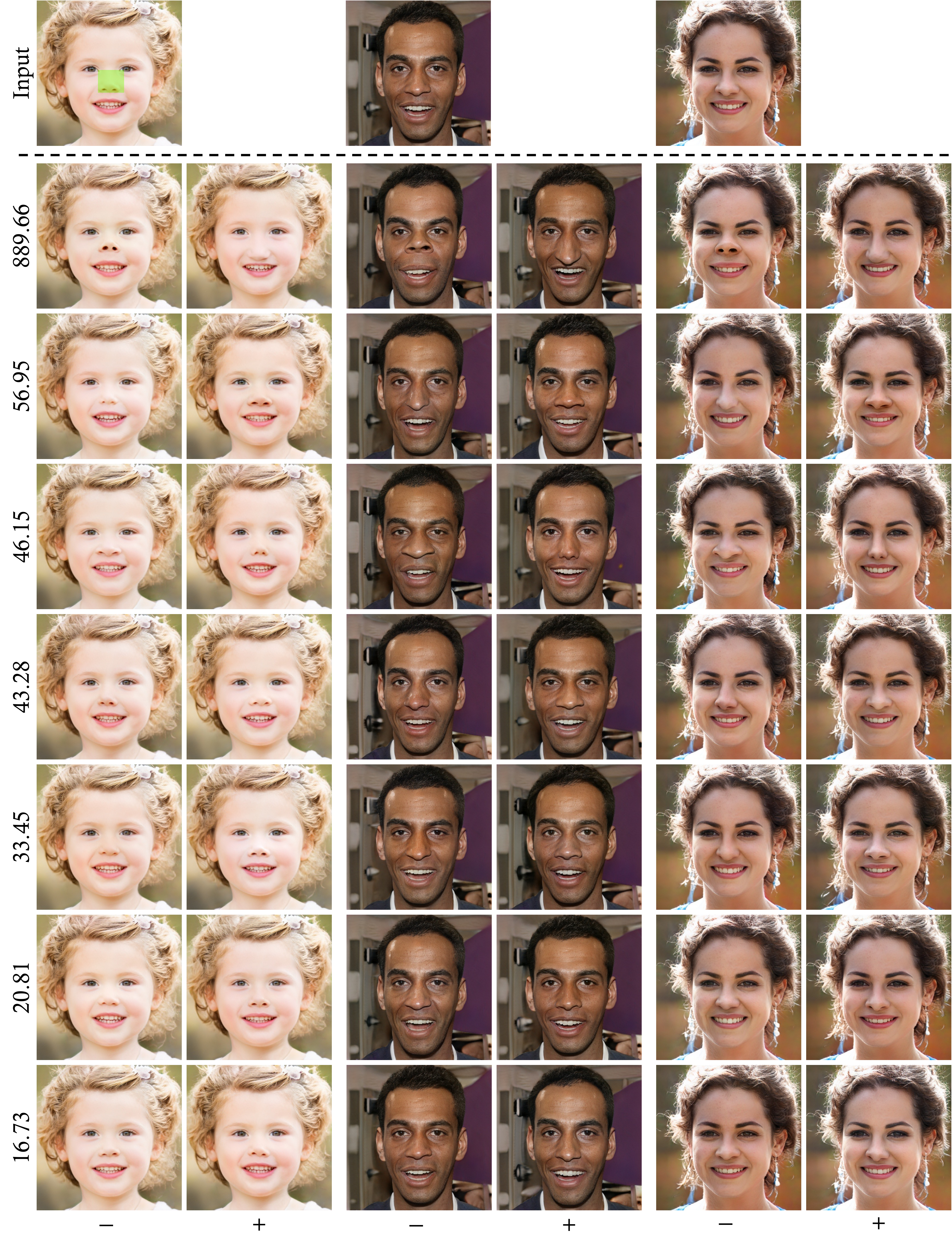}
\vspace{-15pt}
\caption{Visualization results of the first seven principal directions on the nose region of the human faces. The numbers asides the pictures are the eigenvalues corresponding to each direction. The \textbf{\textcolor{green}{green}} mask on the top left image is the region of interest used to compute the directions.}
\label{fig-appendix:face-nose-appendix}
\vspace{-5pt}
\end{figure*}

\begin{figure*}[!ht]
\centering
\includegraphics[width=0.95\textwidth]{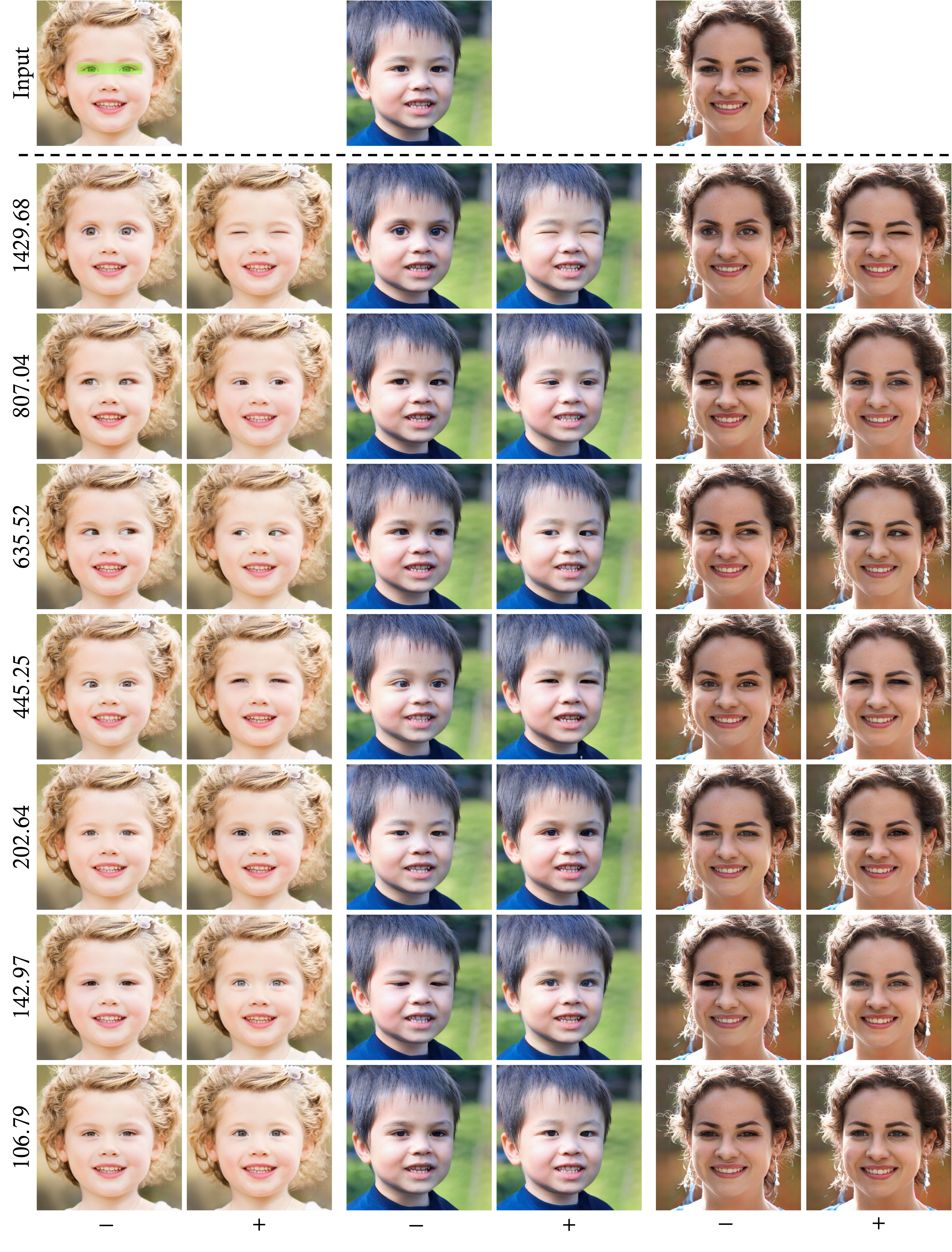}
\vspace{-15pt}
\caption{Visualization results of the first seven principal directions on the eyes region of the human faces. The numbers asides the pictures are the eigenvalues corresponding to each direction. The \textbf{\textcolor{green}{green}} mask on the top left image is the region of interest used to compute the directions.}
\label{fig-appendix:face-eyes-appendix}
\vspace{-5pt}
\end{figure*}

\begin{figure*}[!ht]
\centering
\includegraphics[width=0.95\textwidth]{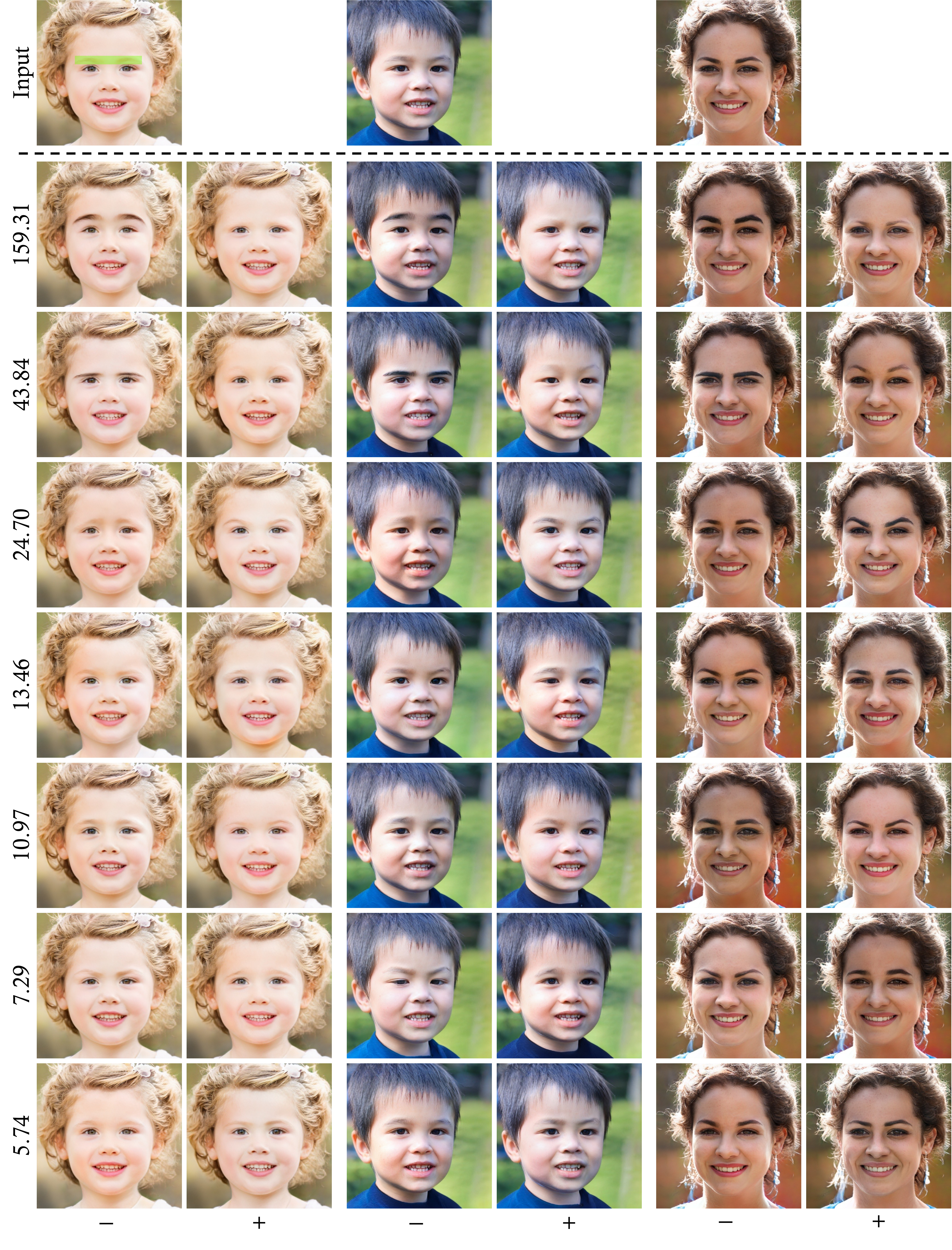}
\vspace{-15pt}
\caption{Visualization results of the first seven principal directions on eyebrows region of the human faces. The numbers asides the pictures are the eigenvalues corresponding to each direction. The \textbf{\textcolor{green}{green}} mask on the top left image is the region of interest used to compute the directions.}
\label{fig-appendix:face-eyebrow-appendix}
\vspace{-5pt}
\end{figure*}

\begin{figure*}[!ht]
\centering
\includegraphics[width=0.95\textwidth]{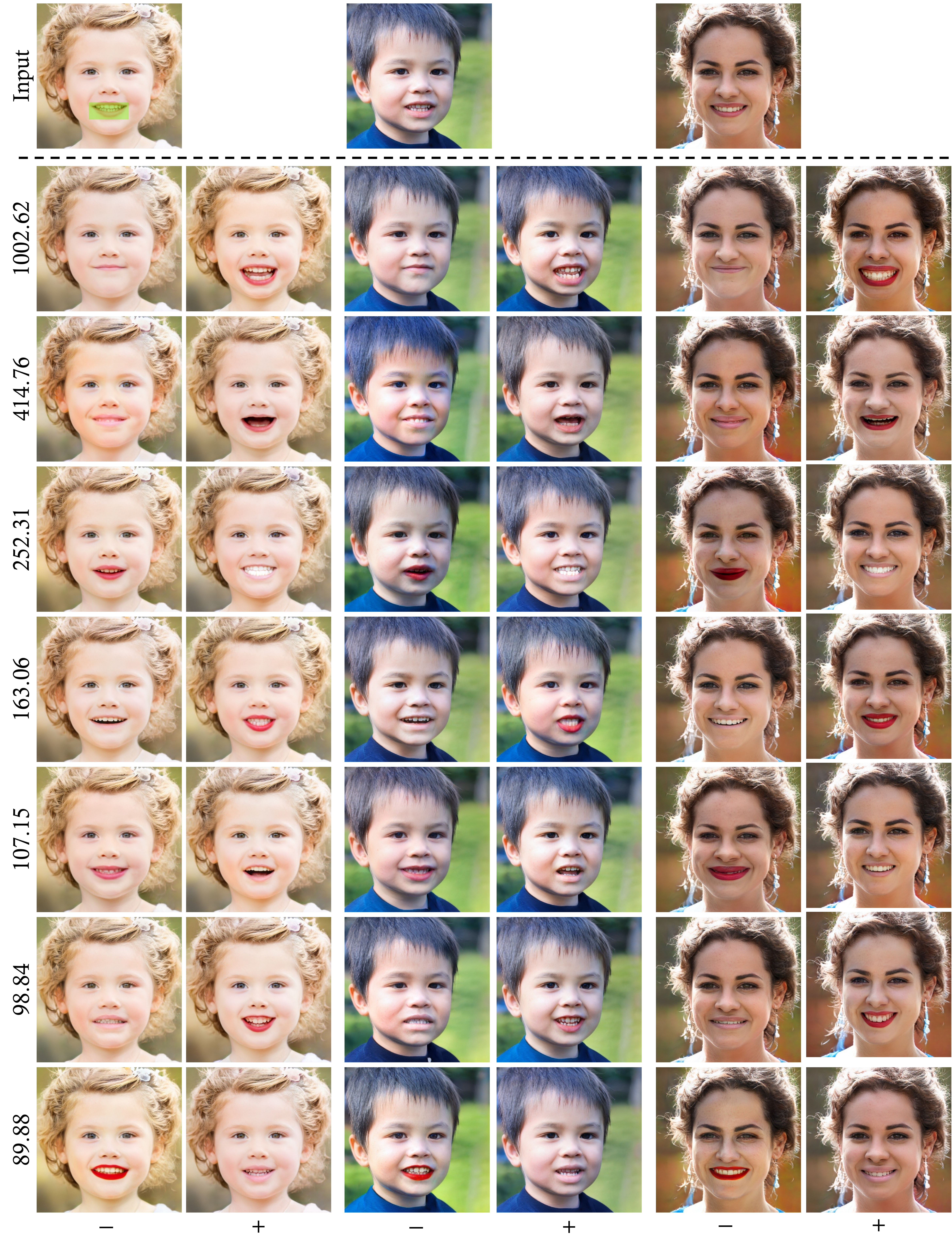}
\vspace{-15pt}
\caption{Visualization results of the first seven principal directions on the mouth region of the human faces. The numbers asides the pictures are the eigenvalues corresponding to each direction. The \textbf{\textcolor{green}{green}} mask on the top left image is the region of interest used to compute the directions.}
\label{fig-appendix:face-mouth-appendix}
\vspace{-5pt}
\end{figure*}

\begin{figure*}[!ht]
\centering
\includegraphics[width=0.95\textwidth]{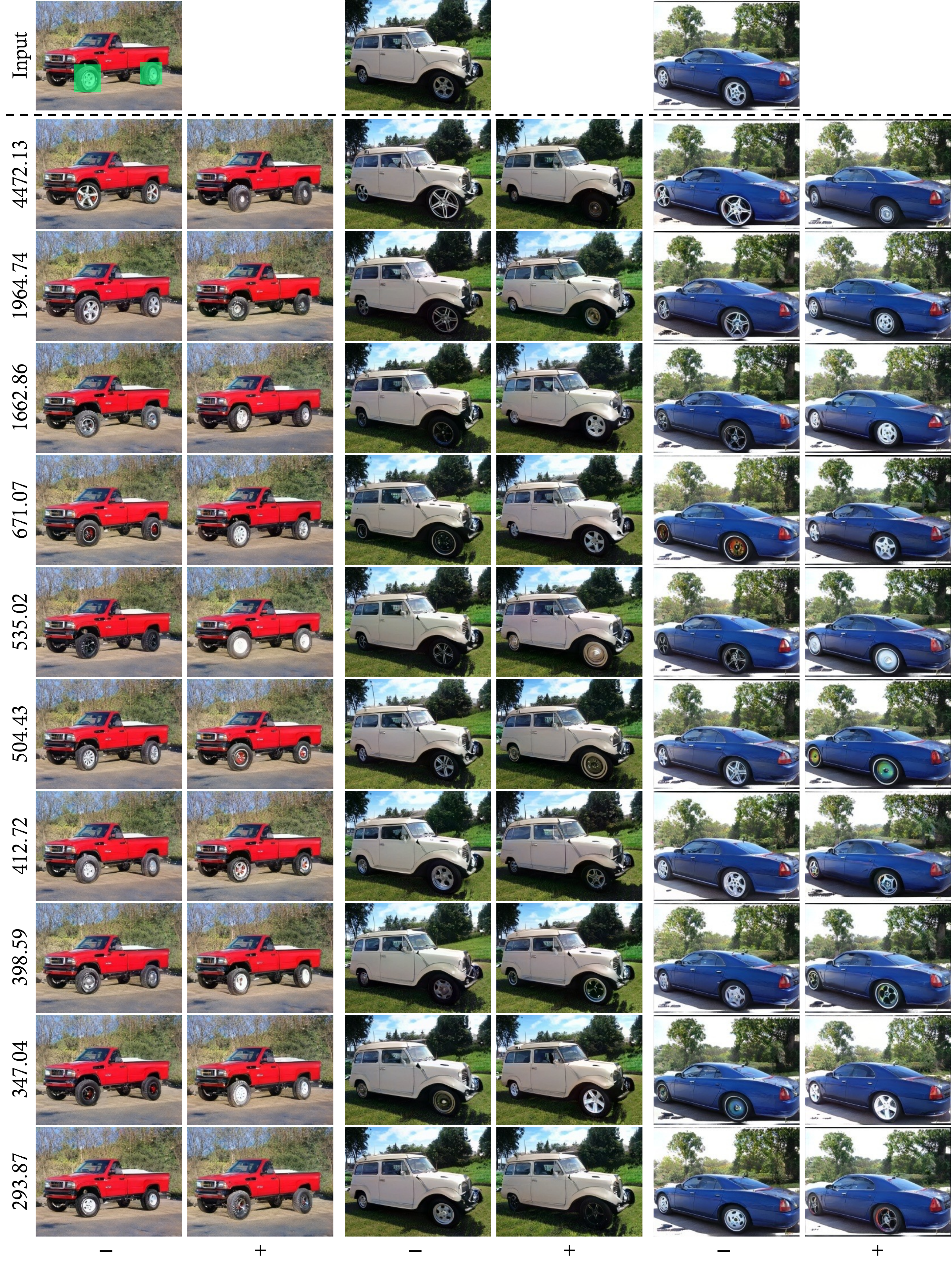}
\vspace{-15pt}
\caption{Visualization results of the first ten principal directions on the wheel region of the car. The numbers asides the pictures are the eigenvalues corresponding to each direction. The \textbf{\textcolor{green}{green}} mask on the top left image is the region of interest used to compute the directions.}
\label{fig-appendix:car-wheel-appendix}
\vspace{-5pt}
\end{figure*}

\begin{figure*}[!ht]
\centering
\includegraphics[width=0.95\textwidth]{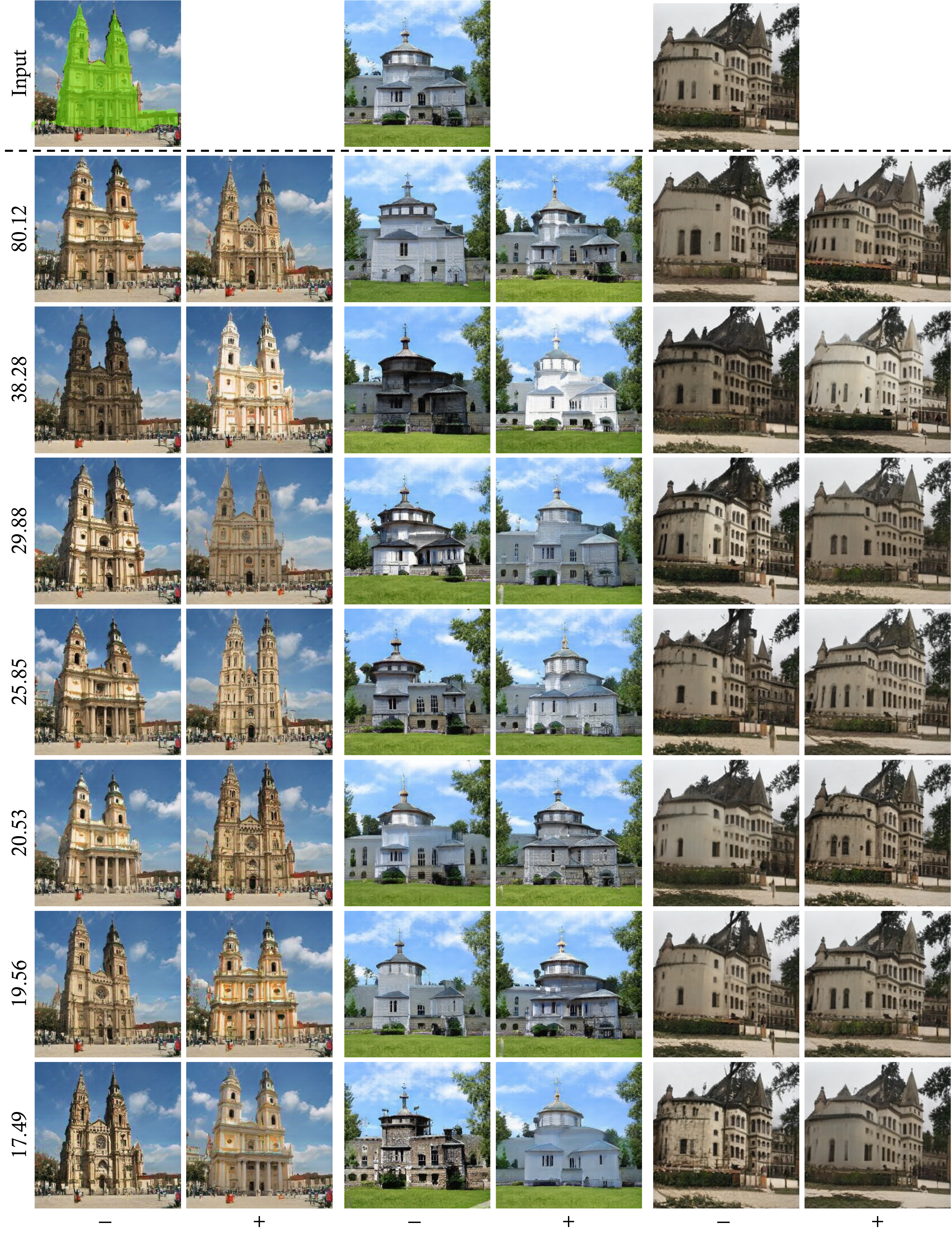}
\vspace{-15pt}
\caption{Visualization results of the first seven principal directions on the church. The numbers asides the pictures are the eigenvalues corresponding to each direction. The \textbf{\textcolor{green}{green}} mask on the top left image is the region of interest used to compute the directions.}
\label{fig-appendix:church-main-appendix}
\vspace{-5pt}
\end{figure*}

\begin{figure*}[!ht]
\centering
\includegraphics[width=0.95\textwidth]{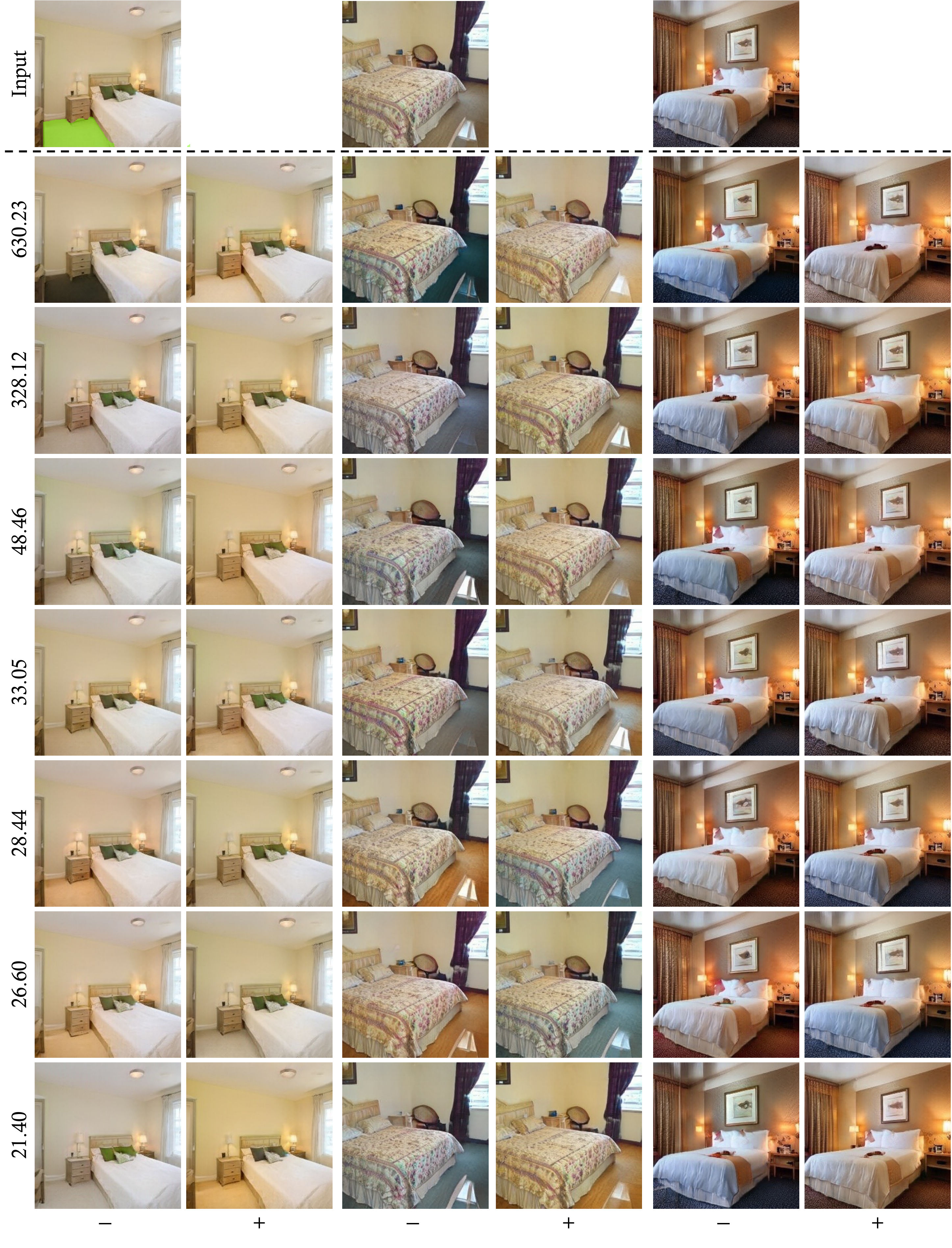}
\vspace{-15pt}
\caption{Visualization results of the first seven principal directions on the floor region of the bedroom. The numbers asides the pictures are the eigenvalues corresponding to each direction. The \textbf{\textcolor{green}{green}} mask on the top left image is the region of interest used to compute the directions.}
\label{fig-appendix:bedroom-floor-appendix}
\vspace{-5pt}
\end{figure*}

\Cref{fig-appendix:scene-car} gives some extra attributes on the  church  and car except for those shown in the main paper.
\Cref{fig-appendix:compare} shows the comparison results with the baselines on wearing lipstick and changing the nose size.
Recall that our attribute vectors are found by solving an eigen-decomposition problem.
Namely, the eigenvectors corresponding to larger eigenvalues are the attribute vectors we need.
Thus, there exists a bunch of meaningful edits using different eigenvectors.
Here we show some edits using different eigenvectors from the same region.
Specifically, \Cref{fig-appendix:face-nose-appendix}, \Cref{fig-appendix:face-eyes-appendix}, \Cref{fig-appendix:face-eyebrow-appendix}, and \Cref{fig-appendix:face-mouth-appendix} show the various semantics found in one image in the regions of the nose, eyes, eyebrow, and mouth, respectively. 
\Cref{fig-appendix:car-wheel-appendix} shows the editing results on the car wheels, \Cref{fig-appendix:church-main-appendix} shows the editing results on the church, and \Cref{fig-appendix:bedroom-floor-appendix} shows the editing results on the bedroom floor.

%% More editing results on StyleGAN

%% Compare with SOTA

%% Additional results on face

%% Additional results on face

%% Additional results on face

%% Additional results on face

%% Additional results on car

%% Additional results on church

%% Additional results on bedroom

\end{document}